\documentclass{article}

\PassOptionsToPackage{numbers, compress}{natbib}

\usepackage[preprint]{neurips_2021}

\usepackage[utf8]{inputenc} 
\usepackage[T1]{fontenc}    
\usepackage{hyperref}       
\usepackage{url}            
\usepackage{booktabs}       
\usepackage{amsfonts}       
\usepackage{nicefrac}       
\usepackage{microtype}      
\usepackage[table,xcdraw]{xcolor}         

\usepackage{microtype}
\usepackage{graphicx}
\usepackage{subfigure}
\usepackage{todonotes}
\usepackage{hyperref}

\usepackage{amsmath,amsfonts,bm}

\def\eqref#1{equation~\ref{#1}}

\def\1{\bm{1}}

\DeclareMathAlphabet{\mathsfit}{\encodingdefault}{\sfdefault}{m}{sl}
\SetMathAlphabet{\mathsfit}{bold}{\encodingdefault}{\sfdefault}{bx}{n}

\usepackage{url}
\usepackage{amsmath,amssymb,dsfont}
\DeclareMathOperator{\EX}{\mathbb{E}}
\usepackage[ruled,vlined]{algorithm2e}
\usepackage{algorithmic}
\usepackage{nicefrac}       
\usepackage{amsfonts}       
\usepackage{subfigure}
\usepackage{textcomp}
\usepackage{multirow}

\usepackage{times}

\title{Improving black-box optimization in VAE latent space using decoder uncertainty}

\author{%
  Pascal Notin$^{1}$\thanks{Correspondence to pascal.notin@cs.ox.ac.uk}
  \hfill
  José Miguel Hernández-Lobato$^{2}$
  \hfill
  Yarin Gal$^{1}$ \\[1.5em]
  $^1$ OATML, Department of Computer Science, University of Oxford \\
  $^2$ Department of Engineering, University of Cambridge \\
 
}

\begin{document}

\maketitle

\begin{abstract}
Optimization in the latent space of variational autoencoders is a promising approach to generate high-dimensional discrete objects that maximize an expensive black-box property (e.g., drug-likeness in molecular generation, function approximation with arithmetic expressions). However, existing methods lack robustness as they may decide to explore areas of the latent space for which no data was available during training and where the decoder can be unreliable, leading to the generation of unrealistic or invalid objects. We propose to leverage the epistemic uncertainty of the decoder to guide the optimization process. This is not trivial though, as a naive estimation of uncertainty in the high-dimensional and structured settings we consider would result in high estimator variance. To solve this problem, we introduce an importance sampling-based estimator that provides more robust estimates of epistemic uncertainty. Our uncertainty-guided optimization approach does not require modifications of the model architecture nor the training process. It produces samples with a better trade-off between black-box objective and validity of the generated samples, sometimes improving both simultaneously. We illustrate these advantages across several experimental settings in digit generation, arithmetic expression approximation and molecule generation for drug design.

\end{abstract}

\section{Introduction}
\label{Sec1_Intro}

We consider the task of optimizing an expensive black-box objective function taking inputs in a \emph{high-dimensional discrete} space. This could be for example finding new molecules for drug design, or automatically generating a computer program that matches a desired output. Solving this task directly in the original space (e.g., with discrete local search methods such as genetic algorithms) may be challenging given the complex structure and high dimensionality of the data.
Recently, Variational autoencoders (VAEs) \citep{kingma2014autoencoding,rezende2014stochastic} have been successfully leveraged to model a wide range of discrete data modalities --- from natural language \citep{bowman2016generating}, to arithmetic expressions \citep{kusner2017grammar}, computer programs \citep{dai2018syntaxdirected} or molecules \citep{Gomez_Bombarelli_2018}. By learning a lower-dimensional continuous representation of objects in their latent space, VAEs allow to transform the original discrete optimization problem into a simpler \emph{continuous} optimization one in latent space. For example, this can be achieved via Bayesian Optimization in the latent space, or via gradient ascent with a jointly-trained neural network predicting the black box property from the latent space representation \citep{Gomez_Bombarelli_2018, bradshaw2019model}.
Initial methods in this area have suffered from the fact that the search in latent space may explore areas for which no data was available at train time, and therefore where the decoder network of the VAE will be unreliable \citep{janz2017actively}: seemingly good candidate points in latent space may be decoded into objects that are invalid, unrealistic or low quality.

\begin{figure*}[h]
    \centering
    \includegraphics[width=\textwidth]{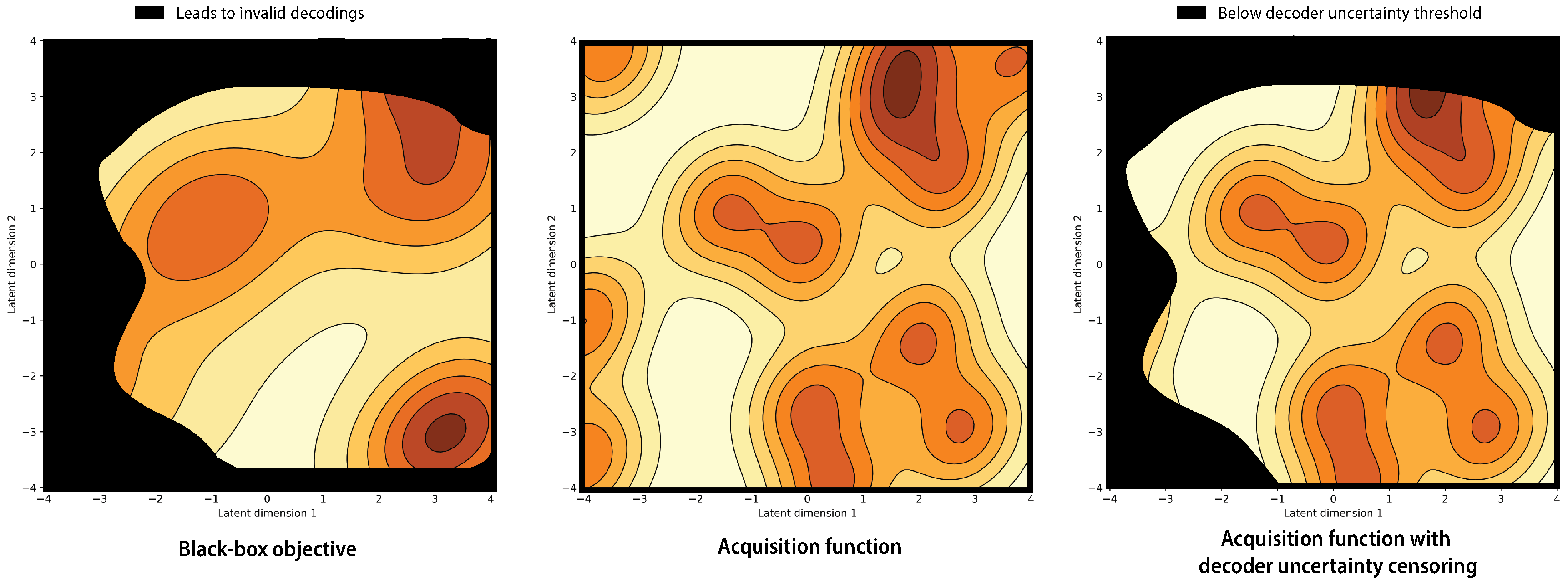}
    \caption{\textbf{Uncertainty-guided optimization in VAE latent space} The goal of black-box optimization in latent space is to attain regions with high values of the back-box objective after decoding, while avoiding the regions that lead to invalid decodings (left). Standard Bayesian Optimization in latent space may query these suboptimal areas (e.g., regions on left hand side, center). High decoder uncertainty regions overlap with regions leading to invalid decodings (right), so that censoring high uncertainty points helps guiding the optimization towards the most promising latent points.}
    \label{Sec1_Fig1}
\end{figure*}

While several methods have been introduced to promote validity of decoded objects (\S\ref{Sec2_Background_Optim}), they either focus on modifying the generative model learning procedure or adapting the decoder architecture to satisfy the syntactic requirements of the data modality of interest. We propose instead to quantify and leverage the uncertainty of the decoder network to guide the optimization process in latent space (Fig.~\ref{Sec1_Fig1}). This approach does not require any change to the model training nor architecture, and can easily be integrated within several optimization frameworks. It results in a better trade-off between the values of the black-box objective and the validity of the newly generated objects, sometimes improving both simultaneously.

To be effective, this method requires robust estimates of model uncertainty for high dimensional structured data. Existing methods for uncertainty estimation in this domain often rely on heuristics or make independence assumptions to make computations tractable (\S\ref{Sec2_Background_Uncertainty}). We demonstrate that such assumptions are not appropriate in our setting, and propose new methods for uncertainty estimation in high dimensional structured data instead. 

Our contributions are as follows:
\begin{itemize}
\item We introduce an algorithm to quantify the uncertainty of high-dimensional discrete data, and use it to estimate the uncertainty of the decoder (\S\ref{Sec3_Estimator});
\item We show how the uncertainty of the decoder can be incorporated across several optimization frameworks, including gradient ascent and Bayesian optimization (\S\ref{Sec4_Uncertainty_guided_Optimization});
\item We illustrate our approach in a digit generation setting --- a simple setup to provide intuition for the method --- then quantify its benefits in the more complex tasks of arithmetic expressions approximation and molecule generation, covering a diverse set of decoder architectures across experiments (Convolutional, Recurrent and Graph Neural Networks) (\S\ref{Sec5_Experiments}).
\end{itemize}
\section{Background}
\label{Sec2_Background}

\subsection{Optimization of high-dimensional discrete objects with generative models}
\label{Sec2_Background_Optim}

Focusing on molecular generation, \citet{Gomez_Bombarelli_2018} propose to train a VAE to learn a distribution over the so-called SMILES representation of molecules (ie. linear sequences of characters) \citep{weininger88smiles}, and subsequently perform the optimization in the latent space. 
Since the SMILES representation follows strict syntactic requirements that are not explicitly enforced by the generative model, promising points in latent space may be decoded into invalid molecules. 
To improve the validity of decoded sequences, \citet{kusner2017grammar} and \citet{dai2018syntaxdirected} develop task-specific grammar rules into the VAE decoder, focusing on use cases in molecular and computer program generation. However, crafting the corresponding rules requires domain-specific knowledge, needs to be designed from scratch for each new task, and may not be straightforward to elicit in the first place (e.g., digit generation example in \ref{Sec5.1_MNIST}). \citet{liu2020chanceconstrained} propose instead to formulate the problem as a chance-constrained optimization task to simultaneously optimize the target property as well as the probability to generate valid sequences. 
A different line of research has focused on representing high-dimensional structured objects as graphs instead \citep{duvenaud2015convolutional,li2018learning}. The Junction Tree VAE (JT-VAE) \citep{jin2019junction} generates systematically valid molecular graphs, by first generating a tree-structured scaffold over a finite set of molecular clusters, and then assembling these clusters back into molecules with a message passing network. The MolDQN \citep{Zhou_2019MolDQN} casts the optimization problem as a reinforcement learning task (double Q-learning), which allows in turn to more naturally extend to simultaneous optimization of different objectives. GraphAF \citep{shi2020graphaf} combines the strengths of autoregressive and flow-based approaches to efficiently generate realistic and valid molecular graphs.
Lastly, \citet{tripp2020sampleefficient} show that the black-box optimization performance can be further enhanced by iteratively retraining the generative model on the data points selected during optimization, with weights that directly depend on their objective function value.

Our approach deviates from all the above in that it is is representation-agnostic (works with sequences or graphs), does not change the model architecture nor the learning procedure and can be combined with several of these approaches to reach even stronger optimization performance.

\subsection{Quantifying model uncertainty}
\label{Sec2_Background_Uncertainty}
Adopting a Bayesian viewpoint, the overall uncertainty of a model in a given region of the input space can be broken down into two types of uncertainty \citep{kendall2017uncertainties}:
\begin{itemize}
    \item \textbf{Epistemic uncertainty:} Uncertainty due to lack of knowledge about that particular region of the input space --- the posterior predictive distribution is broad in that region due to lack of information that can be reduced by collecting more data;
    \item \textbf{Aleatoric uncertainty:} Uncertainty due to inherent stochasticity/noise in the observations in that region --- collecting additional data would not further reduce that uncertainty.
\end{itemize}

We denote input points as $x$, outputs as $y$ and the training data as $\mathcal{D}$. The total uncertainty $\mathcal{U}$ of a model at an input point $x$ is typically measured by the predictive entropy, ie. the entropy of the predictive posterior distribution $P(y|x,\mathcal{D})$:
\begin{equation}
\mathcal{U}(x) = \mathcal{H}(P(y|x,\mathcal{D})) = \sum_{y} - P(y|x,\mathcal{D}) \log P(y|x,\mathcal{D}) dy.
\end{equation}

Denoting $P(\theta|\mathcal{D})$ the posterior distribution over model parameters $\theta$, we can further decompose the predictive entropy $\mathcal{U}$ as the sum of two terms:
\begin{align}
\mathcal{U}(x) = \underbrace{(\mathcal{H}(P(y|x,\mathcal{D})) - \EX_{P(\theta|\mathcal{D})}(\mathcal{H}(P(y|x,\theta)))}_\text{Mutual Information $\mathcal{M}$} +   \underbrace{\EX_{P(\theta|\mathcal{D})}(\mathcal{H}(P(y|x,\theta)))}_\text{Expected entropy $\mathcal{E}$}
\end{align}
The first term --- the Mutual Information $\mathcal{M}$ between model parameters $\theta$ and the prediction $y$ --- is a measure of epistemic uncertainty, as it quantifies the magnitude of the change in model parameters that would result from observing $y$. If the model is uncertain about its prediction for $y$, the change in model coefficients from observing $y$ should be high. Conversely, if the model is confident about its prediction for $y$, model parameters will not vary from observing $y$:
\begin{equation}
\label{Sec2_MI_equation}
\mathcal{M}(x) = \mathcal{H}(P(y|x,\mathcal{D})) - \EX_{P(\theta|\mathcal{D})}(\mathcal{H}(P(y|x,\theta))).
\end{equation}
The second term --- the Expected Entropy $\mathcal{E}$ --- is a measure of the residual uncertainty, ie. the aleatoric uncertainty:
\begin{equation}
\mathcal{E}(x) = \EX_{P(\theta|\mathcal{D})}(\mathcal{H}(P(y|x,\theta))).
\end{equation}
In high dimensional settings, an exact estimation of these different quantities is not tractable, therefore, several approximations and heuristics have been introduced.
The softmax variance, i.e. the variance of predictions across model parameters, has been shown to approximate epistemic uncertainty well in certain settings \citep{carlini2016evaluating, feinman2017detecting,smith2018understanding}. 

In the context of sequential data, the inherent structure in the data generating process often introduces strong dependencies between the output dimensions, e.g., the tokens in a generated sentence. In cases where there exist weak correlations between tokens, quantifying the different types of uncertainties above can be made tractable by ignoring these dependencies \citep{malinin2020uncertainty}, in which case the predictive entropy for a sequence $y = (y_1, y_2, ..., y_L)$ may be approximated as the sum of token-level predictive entropies over the $L$ tokens:
\begin{equation} 
\mathcal{U}(x)
= \sum_{l=1}^L \mathbb{E}_{P(y|x,\mathcal{D})}[\log P(y_l|x,y_{k<l},\mathcal{D})]  \approx \sum_{l=1}^L \mathbb{E}_{P(y_l|x,y_{k<l},\mathcal{D})}[\log P(y_l|x,y_{k<l},\mathcal{D})] .
\label{Sec2_Equation_Token_level_MI}
\end{equation}

Unlike the standard expectation definition which integrates over all $y$, here only $y_l$ is integrated over, and we condition on $y_{k<l}$, which are obtained from a sample from $P(y|x,\mathcal{D})$. The process is repeated for several of these samples and an average is finally computed to reduce variance.

While the above has been shown to work well in certain experiments \citep{malinin2020uncertainty}, valuable information is being discarded when we ignore dependencies across tokens. These dependencies are likely to be informative in applications such as the ones considered in \S\ref{Sec5_CVAE}. Alternative approaches have been suggested which make use of domain-specific metrics in Natural Language Processing such as the BLEU score \citep{Xiao2019Wat}, but these are difficult to extend to other domains. Lastly, a naive Monte Carlo estimation is expected to perform poorly in high dimensions since the majority of samples will be in regions with negligible contribution to the sum (Appendix~\ref{Appendix_Uncertainty_estimator}).
\section{Importance sampling estimator}
\label{Sec3_Estimator}

We consider discrete output points $y$ that belong to a high-dimensional structured object space $\mathcal{S}$ (e.g., long sequences, large graphs), of cardinality $|\mathcal{S}|$. An exact estimation of the Mutual Information between outcomes $y$ and model parameters $\theta$ (equation ~\ref{Sec2_MI_equation}) is impractical because the expectation involves a sum over exponentially many possible outcomes for $y \in \mathcal{S}$. 
 
In lieu of the heuristics previously discussed, we obtain a principled approximation to the Mutual Information via Monte Carlo estimation using \emph{importance sampling}, with an adequately chosen importance distribution.

We denote $q(\theta) \approx P(\theta|\mathcal{D})$ the learnt approximation to the posterior, and assume we can approximate expectations over model parameters by sampling $M$ independent samples from $q(\theta)$.
We can then re-write the Mutual Information $\mathcal{M}$ in equation ~\ref{Sec2_MI_equation} as follows:
\vspace{-2mm}
\begin{equation}
\mathcal{M}(x)
 \approx - \sum_{s=1}^{|\mathcal{S}|}p_s \log p_s + \frac{1}{M}\sum_{m=1}^M{\sum_{s=1}^{|\mathcal{S}|}{p_{s,m} \log p_{s,m}}}
 =  \sum_{s=1}^{|\mathcal{S}|}\underbrace{\left[\frac{1}{M}\sum_{m=1}^M{p_{s,m} \log p_{s,m}} - p_s \log p_s \right]}_{h(y_s)}
 \label{Equation_MI_pre_IS}
\end{equation}
\vspace{-3mm}
\\
where $p_s$ and $p_{s,m}$ are shorthands, respectively, for $P(y=y_s|x,\mathcal{D})$ --- the posterior predictive distribution --- and $P(y=y_s|x,\theta=\theta_m)$ --- the probability of a given output $y_s\in\mathcal{S}$ given $x$ and a sample $\theta_m$ from the approximate posterior distribution over model parameters $q(\theta)$. 

We can then obtain a tractable approximation to equation~\ref{Equation_MI_pre_IS} via importance sampling:
\vspace{-2mm}
\begin{equation}
\mathcal{M}(x) = \sum_{s=1}^{|\mathcal{S}|} h(y_s) \cdot \frac{1}{\bar{p}(y_s)} \cdot \bar{p}(y_s) = \mathbb{E}_{\bar{p}}{\left[h(y) \cdot \frac{1}{\bar{p}(y)}\right]}
\approx \frac{1}{N} \sum_{s=1}^N {\left[h(\tilde{y}_s) \cdot \frac{1}{\bar{p}(\tilde{y}_s)}\right]}
\end{equation}
\label{Sec3_Equation_IS_MI}
with $\tilde{y}_s \sim \bar{p}(.)$, where $\bar{p}$ is the importance distribution.

We choose the importance distribution to be the approximate posterior predictive defined over the outputs. We generate an outcome $\tilde{y}_s$ by first sampling a set of parameters $\tilde{\theta}_{0}$ from the approximate posterior, and then generating $\tilde{y}_s$ from a model defined by that set of parameters $\tilde{\theta}_{0}$.
This distribution will sample mostly from regions in $\mathcal{S}$ with high probability under the true posterior predictive for input $x$. This is in contrast to a naive sum over all possible outcomes $y$, many of which will have a negligible contribution to the sum. This gives rise to an estimator of Mutual information (obtained with Algorithm 1) with lower variance than its naive Monte Carlo counterpart (see Appendix~\ref{Appendix_Uncertainty_estimator}).

\begin{algorithm}[h]
\label{Sec3_Algorithm_IS-MI}
\begin{algorithmic}
\caption{Importance sampling estimator of MI} 
    \FOR{$s=1$ {\bfseries to} $N$}
	    \STATE Sample\ $\tilde{\theta}_{0} \sim q(\theta)$ ; $\tilde{y}_s \sim P(y|x,\theta=\tilde{\theta}_{0})$ ; 
	    \FOR{$m=1$ {\bfseries to} $M$}
	        \STATE Sample\ $\tilde{\theta}_m \sim q(\theta)$ ; Compute\ $p_{s,m}$ = $P(y=\tilde{y}_s|z,\theta=\tilde{\theta}_m)$ ;
		\ENDFOR
	    \STATE Compute $p_s = \frac{1}{M}\sum_{m=1}^M{p_{s,m}}$; $h_s = \frac{1}{M}\sum_{m=1}^M({p_{s,m} \log p_{s,m})} - p_s \log p_s$ ;
	\ENDFOR
	\STATE Return $\mathcal{M}(x)$ = $\frac{1}{N} \sum_{s=1}^{N} \left[h_s \cdot \frac{1}{p_s} \right]$ \\
\end{algorithmic}
\end{algorithm}
\section{Uncertainty-guided optimization in VAE latent space}
\label{Sec4_Uncertainty_guided_Optimization}

\textbf{Black box optimization in VAE latent space.} We want to optimize the black-box objective $\mathcal{O}$ over a high dimensional discrete object space $\mathcal{S}$. We train a VAE, with encoder $g$ and decoder $f$, to learn a continuous lower-dimensional embedding of objects in $\mathcal{S}$. The optimization of $\mathcal{O}$ is then performed in latent space and the best candidates are subsequently decoded into the original space. As discussed in \S~\ref{Sec2_Background_Optim}, this may lead to invalid or unrealistic decodings when the decoder $f$ operates in regions different from the ones seen during training. We propose to detect this regime by quantifying the epistemic uncertainty of the decoder: avoiding regions with high epistemic uncertainty for the decoder will make the overall optimization process more efficient by avoiding invalid decodings. 
We next cover two optimization approaches commonly used in latent space optimization settings and discuss how we can leverage the uncertainty of the decoder to guide the optimization process.

\textbf{Bayesian Optimization with an uncertainty-aware surrogate model or uncertainty censoring.} We first train a surrogate model, e.g., a Gaussian Process \citep{Mchutchon2011GP}, to predict $\mathcal{O}(x)$ based on its latent representation $z$. We then perform Bayesian Optimization using an appropriate acquisition function (e.g., Upper Confidence Bound or Expected Improvement heuristic). There are two main ways to incorporate the decoder uncertainty to guide this process. The first approach consists in training the surrogate model on an objective that penalizes points with high uncertainty (e.g., optimizing $\mathcal{O}(x) - \alpha \cdot \mathcal{M}(z)$). Another method is to censor proposal points $z$ that would have a Mututal Information $\mathcal{M}(z)$ above a predefined uncertainty threshold $\mathcal{T}$ (e.g., highest value observed on the training data) at each step of a batch Bayesian Optimization process. 

\textbf{Uncertainty-constrained gradient ascent.} A common architecture design when performing black-box optimization in latent space is to jointly train the VAE with an auxiliary network $h$ (Fig.\ref{Appendix_Figure_Joint_training_architecture}) that predicts the value of the black box objective $\mathcal{O}(x)$ from the encoding $z$ of $x$ in latent space \citep{Gomez_Bombarelli_2018,bradshaw2019model,jin2019junction}. This construct is particularly useful in constrained optimization settings in which we want to perform a local search in latent space to maximize $\mathcal{O}$ while remaining close to a known input object.
The joint training consists of optimizing the sum of the VAE loss (i.e., the ELBO) and the loss from the black-box objective prediction (e.g., MSE loss for a continuous output $\mathcal{O}$) via gradient descent, backpropagating gradients through the entire architecture.
To optimize objects under this framework, we start from a set of initial points in latent space $z$ --- either a random sample of latent points, or a subset of points $x$ that we encode in the latent space ($z = g(x)$). We then compute the gradient $\nabla_{z}{h}$ of the auxiliary network with respect to $z$ and perform gradient ascent $ z \leftarrow z + \alpha \cdot \nabla_{z}{h}$. We repeat this process a few times until satisfying a stopping criteria (e.g., threshold on predicted values $h(z)$ or after a fixed number of gradient updates). Finally, we decode the latest latent positions to obtain the set of candidates $\Tilde{x} = f(z)$ and measure their actual properties $\mathcal{O}(\Tilde{x})$.
We can further improve the quality of the candidate set by censoring the moves in latent during gradient ascent that would result in a value of uncertainty above a predefined threshold $\mathcal{T}$.

\section{Experimental results}
\label{Sec5_Experiments}

After describing the common experimental setting across applications, we demonstrate the effectiveness of using the uncertainty of the decoder to guide the optimization for constrained digit generation. Our objective is to illustrate the concepts introduced above in a simple and intuitive example. We then move on to quantify the benefits of our approach in the more complex cases of arithmetic expression approximation and molecular generation for drug design. 

\textbf{Uncertainty estimators and baselines} Across all experiments, we quantify the Mutual Information between outputs and decoder parameters with both the Importance sampling estimator (IS-MI) described in \S\ref{Sec3_Estimator} and based on the token independence approximation (TI-MI) described in \S\ref{Sec2_Background_Uncertainty}. Sampling from model parameters is achieved via Monte Carlo dropout \citep{gal2016dropout}. We compare optimization results with two baselines: the standard approach that fully ignores decoder uncertainty, and an approach in which we censor proposal points with low probability under the prior distribution (standard normal) of the VAEs in latent space (referred to as `NLLP').

\textbf{Optimization} We perform uncertainty-guided optimization in latent space as per the two approaches described in \S\ref{Sec4_Uncertainty_guided_Optimization}. For Bayesian Optimization, we train a single task Gaussian Process as our surrogate model based on a random subset of training points embedded in latent space and their corresponding black-box objective values. We then perform several iterations of batch Bayesian Optimization using the Expected Improvement heuristic as our acquisition function. At each iteration we select a batch of 20 latent vectors by sequentially maximizing the acquisition function. We select the point with the highest predicted target value for which the decoder uncertainty is below a predefined threshold (e.g., $99^{th}$ percentile of decoder uncertainty values observed on the training set) or the one with lowest uncertainty if no point in the batch is below the threshold. We re-train the surrogate model with the newly generated point at each step.
For gradient ascent, we randomly sample points from the training set, embed them in latent space, and use these as our starting positions. We then compute the gradient of the auxiliary property network with respect to latent positions and accept proposal moves along these directions if the decoder uncertainty at the corresponding position in latent is below a predefined threshold (e.g., $99^{th}$ percentile of decoder uncertainty on the training set).
All optimization experiments reported below are carried out 10 times independently with different random seeds. 

\subsection{Illustrative example in digit generation}
\label{Sec5.1_MNIST}

\textbf{Setup} In this first setting, our objective is to generate \emph{valid} images of the digit 3 that are as \emph{thick} as possible. We train a VAE model generating images of the digit 3 jointly with an auxiliary network predicting their thickness. We use a `Conv-Deconv'\citep{Higgins2017} architecture for the VAE and a 3-layer feedforward network for property prediction. The underlying data consist of grayscale images of the digit 3  extracted from the MNIST dataset \citep{Lecun1998MNIST} that we discretize to form tensors of binary values. We use the sum of pixel intensities across a given image as a proxy for the thickness of the corresponding digit. An unconstrained optimization in the latent space would ultimately lead to the generation of \emph{invalid} white `blobs'. To avoid this failure mode and promote validity of the resulting candidate set, we leverage the uncertainty of the decoder network. In order to assess the validity of objects, we independently train a deep Convolutional Neural Network to classify images of the digit 3  (binary classification).

\begin{figure}[ht]
\begin{minipage}[m]{0.48\linewidth}
\vspace{0pt}
\centering
    \includegraphics[scale=0.55]{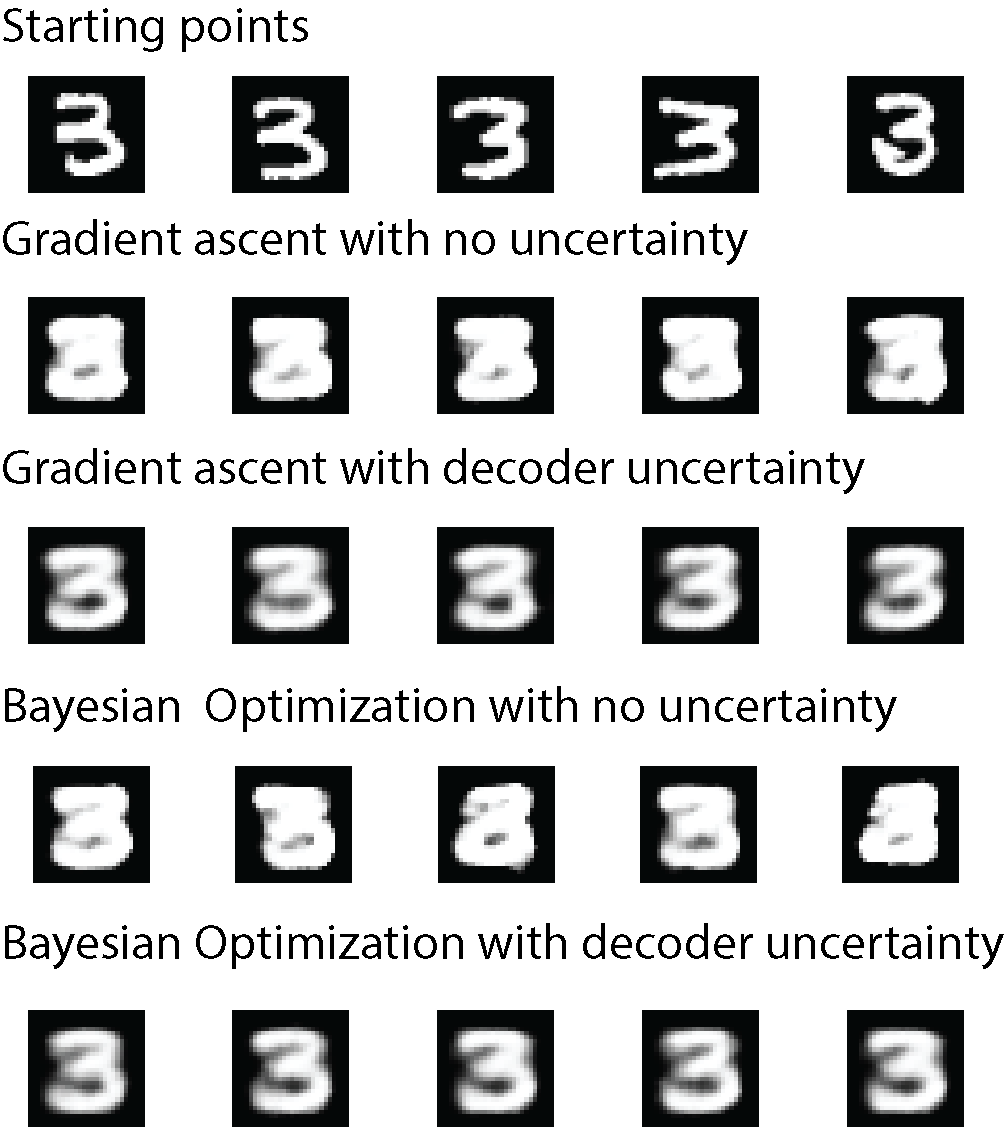}
    \caption{\textbf{Top 5 decoded digits after optimization} Leveraging decoder uncertainty helps preventing the generation of invalid digits.}
    \label{Sec5_Fig_MNIST_Top_images}
\end{minipage}
\hfill
\begin{minipage}[m]{0.48\linewidth}
\vspace{0pt}\raggedright
\centering
\resizebox{.98\textwidth}{!}{
\begin{tabular}{ccc}
\toprule
\textbf{Method} & \textbf{Top 10 avg.} $\uparrow$ & \textbf{Validity (\%)} $\uparrow$ \\
\toprule
No uncertainty & $-0.96\pm 0.04$ & $77\% \pm 0.6\%$\\
NLLP & $-0.97 \pm 0.05$& $76\%\pm 0.8\%$ \\
TI-MI & $-0.72\pm 0.06$ & $96 \% \pm 0.5\%$\\
IS-MI & $\textbf{-0.70}\pm \textbf{0.04}$ & $\textbf{98\%} \pm \textbf{0.5\%}$\\
\bottomrule
\end{tabular}

}
\caption{\textbf{Arithmetic expressions approximation results} Mutual Information-based constraints during Bayesian optimization in latent space help promoting higher validity \% of decodings while increasing black-box objective values. Uncertainty threshold values used for censoring candidate points are based on decoder uncertainty values observed on the training data ($95^{th}$ percentile).}
\label{Sec5.2_Arithmetic_results}
\end{minipage}
\end{figure}

\textbf{Results} Latent points with high decoder uncertainty lead to a higher rate of invalid decoded digits (Appendix~\ref{Appendix_B.3_Digit_generation_Uncertainty_estimator}), which help avoid the aforementioned failure mode during optimization. 
For both Bayesian Optimization and gradient ascent, we observe that the decoder uncertainty constraints help ensuring the generated digits remain valid, while preserving the ability of the optimization algorithm to increase the thickness (Fig.\ref{Sec5_Fig_MNIST_Top_images}).

\subsection{Arithmetic expressions approximation}

\textbf{Setup} We follow an experimental design similar to \citet{kusner2017grammar}, in which we seek to optimize univariate arithmetic expressions generated by a formal grammar (rules and examples are provided in Appendix~\ref{Appendix_C_Arithmetic_expression}). The objective is to find an expression that minimizes the mean squared error (MSE) with respect to a predefined target expression ($1/3 * x * sin(x*x)$). More specifically, since the presence of exponentials in expressions may results in very large MSE values, the black-box objective is defined as $-log(1+MSE)$. We train a `Character VAE' (CVAE) \citep{kusner2017grammar} on $80,000$ expressions generated by the formal grammar, then perform optimization in the latent space.

\textbf{Results} We observe that methods leveraging the decoder uncertainty result in almost always valid decodings, and reach higher average values of the black box objective for valid decoded expressions compared to baselines (Fig.~\ref{Sec5.2_Arithmetic_results}; gradient ascent results in Appendix~\ref{Appendix_C.4_Arithmetic_expression_Optimization}). In particular, censoring candidate points based on their probability under the standard normal prior does not help promoting validity of decodings at all. In this setting with relatively short sequences (arithmetic expressions have at most 19 characters), leveraging the TI-MI or IS-MI estimators leads to comparable performance.

\subsection{Molecule generation for drug discovery}

Molecular generation for drug design seeks to identify new molecules satisfying desired chemical properties. Molecules are typically either represented as sequences of characters, using their SMILES representation \citep{weininger88smiles}, or as graphs of atoms \citep{duvenaud2015convolutional}. We demonstrate the effectiveness of the approach described in \S~\ref{Sec4_Uncertainty_guided_Optimization} for these two different representations: experiments with the `Character VAE' (CVAE) for molecules \citep{Gomez_Bombarelli_2018} leverage the SMILES representation, while experiments with the JT-VAE \citep{jin2019junction} are based on a graph representation of molecules. For both architectures, we trained our models on a set of 250k drug-like molecules from the ZINC dataset \citep{Irwin2012ZINCAF}.

\subsubsection{Character VAE (CVAE)}
\label{Sec5_CVAE}

\textbf{Setup} We jointly train a CVAE model, which learns to encode and decode molecules SMILES strings, along with an auxiliary property network that predicts a target property of these molecules. Following prior work \citep{kusner2017grammar, dai2018syntaxdirected, jin2019junction}, we define the black-box objective as the octanol-water partition coefficient penalized by the synthetic accessibility score and the number of long cycles, (Appendix~\ref{Appendix_D.1_Molecule_CVAE_Data}) and we refer to it as `Penalized logP' for brevity.
Since the SMILES representation of molecules follows a strict syntax that determines whether a given expression is valid or not, we are interested in generating molecules that simultaneously maximize the target property and represent valid SMILES expressions.

\begin{figure*}[h]
    \centering
    \includegraphics[width=5.5in,scale=1.0]{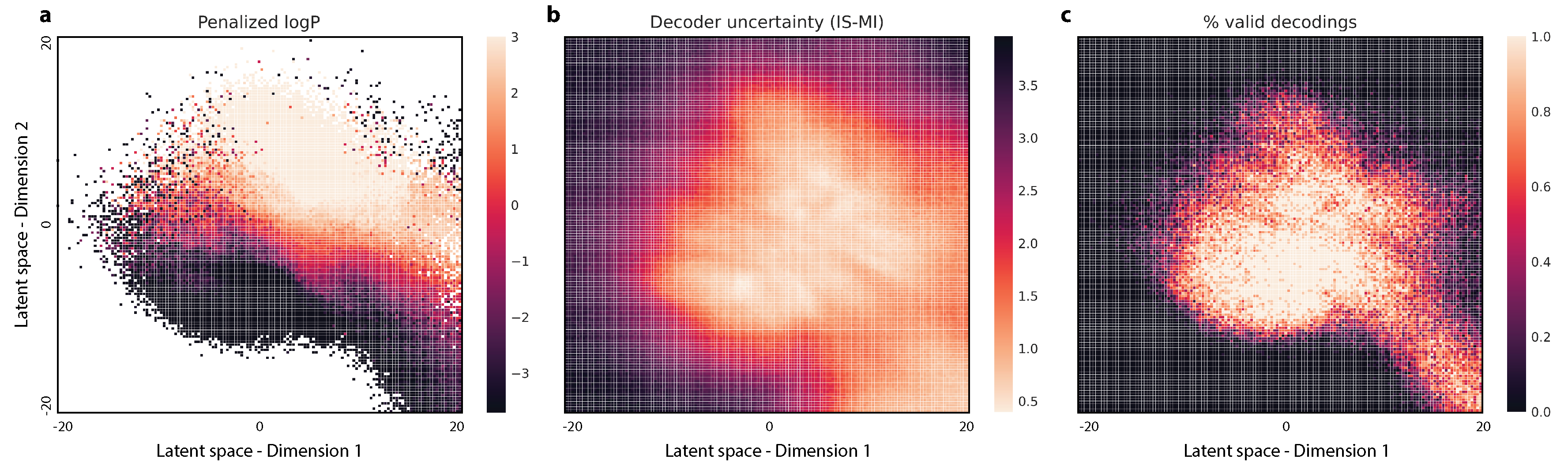}
    \vspace{-1mm}
    \caption{\textbf{CVAE latent space visualization.} We apply Principal Component Analysis on the embedding of the full training data and keep the first 2 components. We then create a grid on the resulting 2D-space and measure the penalized logP (a), decoder uncertainty (b) and the proportion of valid decodings in that region (c) (a \& c averaged over 300 decodings; b measured via IS by sampling 100 times from the importance distribution, and averaging over 100 samples of model parameters.; for a, white squares correspond to regions where none of the 300 decodings are valid). We observe a strong overlap between decoder uncertainty (b) and validity of decodings (c).}
    \label{Fig_CVAE_Latent_space_viz}
\end{figure*}
\vspace{-1mm}

\begin{figure*}[h]
    \centering
    \includegraphics[width=5.5in,scale=1.0]{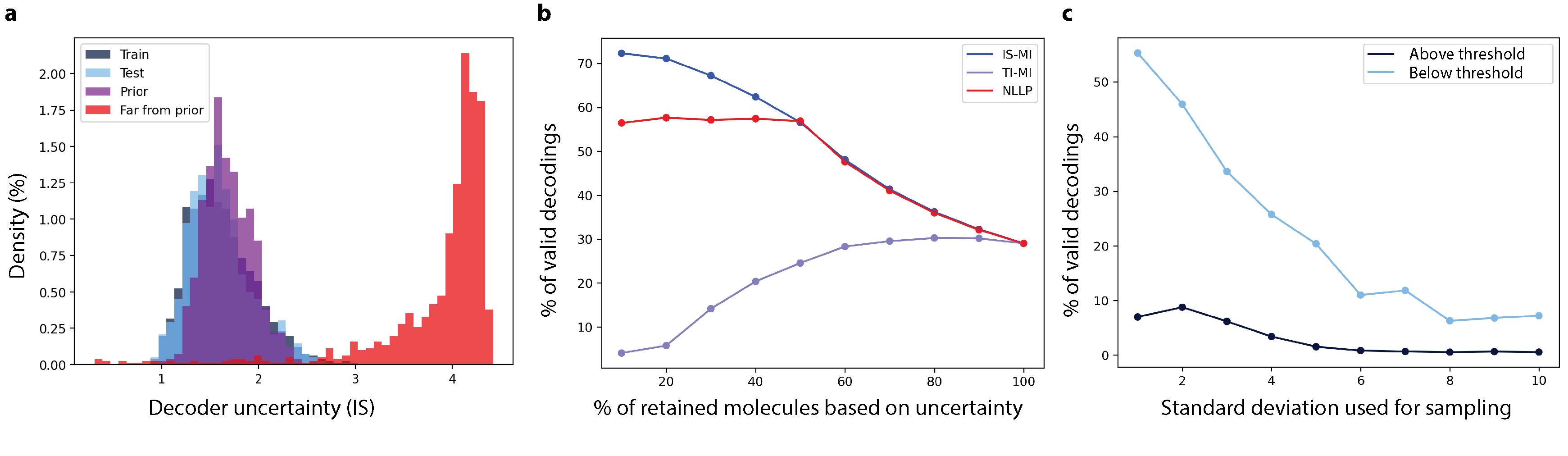}
    \caption{\textbf{Uncertainty estimator.} a) Distribution of decoder uncertainty values (IS-MI) for 1k samples for 4 distinct sets (train \& test set samples embedded in latent space; samples from the prior; samples far from the prior). b) Valid decodings (\%) as a function of the proportion of samples kept based on their uncertainty --- eliminating points with high uncertainty first (dataset comprised of 50\% samples from test set \& 50\% of samples far from the prior). The IS-MI estimator has superior ability to identify points leading to invalid decodings. c)  Valid decodings (\%) for samples from a normal distribution with increasing standard deviation. Samples with decoder uncertainty below a predefined threshold (maximum IS-MI value observed on training data) have a much higher rate of valid decodings. Points above the threshold are very likely to lead to invalid decodings.}
    \label{Fig_CVAE_Uncertainty_estimator}
\vspace{-2mm}
\end{figure*}

\textbf{Results} We first verify that our estimator is able to discriminate points in-distribution (low uncertainty) vs out-of-distribution (high uncertainty). We consider 4 distinct sets of points in latent: embeddings into latent space of a random sample from the train and test sets, random samples from the VAE prior (standard normal) and random samples ``far from the prior'' (we sample from an isotropic gaussian with standard deviation equal to 10). As can be seen on Fig.\ref{Fig_CVAE_Uncertainty_estimator}a, uncertainty estimates for the first 3 sets strongly overlap while being disjoint from the estimates corresponding to points far from the prior.  Furthermore, we observe a strong correlation between low decoder uncertainty and regions that lead to valid SMILES decodings (Fig.~\ref{Fig_CVAE_Latent_space_viz}). This is corroborated by the analysis described in Fig.~\ref{Fig_CVAE_Uncertainty_estimator}c: when considering latent points ``far from the prior'', points for which the decoder uncertainty is lower than a predefined threshold (e.g., maximum value observed on training data) will lead to a significantly higher proportion of valid decoded molecules compared to latent points with uncertainty above the threshold. This is critical as it allows to censor points that will likely lead to invalid decodings, even when we move \emph{far from the prior} in latent space.

For the Bayesian Optimization experiments, we investigate the impact of different bounds on the space we optimize within, as well as different uncertainty thresholds. As we increase the bounds, we typically reach higher optima, at the cost of a higher fraction of invalid decodings during search. We obtain higher validity \% and penalized logP values when leveraging the decoder uncertainty (Table \ref{Table_CVAE_BO_results}). In this setting, the token-level independence assumption (TI-MI) leads to poor performance compared to the importance sampling-based estimator (IS-MI). Results are also robust to the choice of decoder uncertainty thresholds (\ref{Appendix_D.4_Molecule_CVAE_Optimization}).

\begin{table}
\begin{center}
\vspace*{-1mm}
\caption{\textbf{CVAE - Bayesian Optimization results.} Censoring proposal points with high decoder uncertainty values with the IS-MI estimator helps increase validity across experiments. As we increase the bounds on the Bayesian Optimization search space, validity \% generally decreases but remains 5-10x higher when leveraging IS-MI compared to baselines. This is critical as is helps uncover molecules with very high penalized logP values.}
\begin{tabular}{ccccc}
\toprule
\textbf{Search}  & \textbf{Decoder} &  \multicolumn{2}{c}{\textbf{Penalized logP}} & \textbf{Validity}\\
\textbf{bounds} & \textbf{uncertainty} & \textbf{Top 1} $\uparrow$ & \textbf{Avg. top 10} $\uparrow$ & \textbf{(\%)} $\uparrow$ \\
\toprule
5  & None  & $4.0  \pm 0.2$& $2.5  \pm 0.2$& $22\% \pm 1.4\%$\\
   & NLLP  & $4.2  \pm 0.2$& $2.7  \pm 0.1$& $30\% \pm 1.3\%$\\
   & TI-MI & $4.1  \pm 0.3$& $2.3  \pm 0.1$& $21\% \pm 0.8\%$\\
   & IS-MI & $\textbf{4.5}  \pm \textbf{0.2}$& $\textbf{3.0}  \pm \textbf{0.1}$& $\textbf{33\%} \pm \textbf{1.8\%}$\\ 
\midrule
10 & None  & $3.9  \pm 1.2$& $-2.3 \pm 2.8$& $1\%  \pm 0.4\%$\\
   & NLLP  & $2.9  \pm 0.8$& $0.5  \pm 0.8$& $3\%  \pm 0.7\%$\\
   & TI-MI & $5.9  \pm 3.6$& $1.1  \pm 1.5$& $2\%  \pm 0.4\%$\\
   & IS-MI & $\textbf{6.6} \pm \textbf{0.6} $& $\textbf{1.6} \pm \textbf{0.8}$&$ \textbf{11\%} \pm \textbf{0.8\%}$\\ 
\midrule
15 & None  & $10.3 \pm 4.3$& $5.0  \pm 2.6$& $1\%  \pm 0.3\%$\\
   & NLLP  & $3.9  \pm 2.5$& $0.8  \pm 1.2$& $1\%  \pm 0.3\%$\\
   & TI-MI & $6.7  \pm 3.8$& $6.4  \pm 3.9$& $1\%  \pm 0.3\%$\\
   & IS-MI & $\textbf{27.6} \pm \textbf{2.2}$& $\textbf{9.9} \pm \textbf{1.3} $& $\textbf{5\%} \pm \textbf{0.7\%}$\\
\bottomrule
\end{tabular}
\label{Table_CVAE_BO_results}
\end{center}
\vspace{-2mm}
\end{table}

\subsubsection{Junction Tree VAE (JT-VAE)}
\label{Sec5_JTVAE}

\textbf{Setup} We train a Junction Tree VAE model (JT-VAE) \citep{jin2019junction} using the same dataset of 250k molecules (ZINC) and black-box objective (penalized logP) as for the CVAE experiments. 
All molecules generated by the JT-VAE are valid by design. However, not all generated molecules will be of high \emph{quality}, which we assess with the quality filters proposed by \citet{Brown_2019} that aim at ruling out ``compounds which are potentially unstable, reactive, laborious to synthesize, or simply unpleasant to the eye of medicinal chemists.''
We show that it is straightforward to attain state-of-the-art performance in terms of penalized logP values with the basic optimization approaches described in \S~\ref{Sec4_Uncertainty_guided_Optimization} by moving sufficiently `far away' in latent, but that in doing so we tend to generate molecules that never pass quality filters. Factoring in decoder uncertainty during optimization helps generate new molecules with both high penalized logP values and high quality.

Using notations from \S~\ref{Sec3_Estimator}, sampling a new object $\tilde{y}_s$ is achieved by successively decoding from the junction tree decoder and then the graph decoder. We then decompose $\log{p_{s,m}}$ -- the log probability of the sampled graph molecule -- as the sum of the log probabilities corresponding to the different predictions made by the junction tree decoder and graph decoder, namely the topology and node predictions in the junction tree decoder, and the subgraph prediction in the graph decoder. \\
We replicate the analysis described in \ref{Sec5_CVAE} with the 4 distinct datasets in latent space (i.e., train, test, prior and far from prior) and observe similar results: the histogram of decoder uncertainty values for points ``far from the prior'' is disjoint from the other three histograms (Appendix~\ref{Appendix_E.3_Molecule_JTVAE_Uncertainty_estimator}), confirming the ability of the estimator to identify out-of-distribution points.

\textbf{Results} Both gradient ascent (Table~\ref{Sec5_Table_JTVAE_GA_results}) and Bayesian Optimization (Appendix~\ref{Appendix_E.4_Molecule_JTVAE_Optimization}) allow to generate new molecules with state-of-the-art performance in terms of penalized logP (Table~\ref{Appendix_E_Table_Molecular_generation_top_performance}). However, the majority of these molecules do not pass quality filters. Leveraging decoder uncertainty leads to the generation of high logP and high quality molecules. Using likelihood under the prior (NLLP) to achieve the same is detrimental to optimization performance.

\begin{table}
\begin{center}
\vspace{-1mm}
\caption{\textbf{JT-VAE - Gradient ascent results.} We obtain state-of-the-art performance in terms of penalized logP via gradient ascent. However, most generated molecules are of very low quality (only ~1\% pass the quality filters from \citet{Brown_2019}). Leveraging the uncertainty of the decoder (IS-MI) during optimization helps generating molecules with high penalized logP and high quality. NLLP constraints help maintain high quality but lead to suboptimal black-box objective values. Results with different hyperparameters and threshold values for each method are reported in Table~\ref{Appendix_E_Table_JTVAE_GA_results}.}
\resizebox{\textwidth}{!}{
\begin{tabular}{cccccc}
\toprule
\textbf{Decoder} &  \multicolumn{2}{c}{\textbf{Penalized logP - Before filters}} & \textbf{Quality top 10} & \multicolumn{2}{c}{\textbf{Penalized logP - Passing filters}}\\
\textbf{uncertainty} & \textbf{Top 1} $\uparrow$ & \textbf{Avg. top 10} $\uparrow$ & \textbf{(\%)} $\uparrow$ & \textbf{Top 1} $\uparrow$ & \textbf{Avg. top 10} $\uparrow$\\
\toprule
None & $\textbf{23.7} \pm \textbf{1.3}$& $\textbf{17.0} \pm \textbf{0.6}$& $ 1\% \pm 1\%$& $1.2 \pm 1.2$& $0.3 \pm 0.3$ \\
NLLP & $3.0 \pm 0.1$& $2.5 \pm 0.1$& $ 82\% \pm 6\%$& $3.0 \pm 0.1$& $2.0 \pm 0.2$ \\
IS-MI & $8.4 \pm 10.8$& $6.0 \pm 0.3$& $ \textbf{89\%} \pm 3\textbf{\%}$& $\textbf{7.7} \pm \textbf{0.7}$& $\textbf{5.3} \pm \textbf{0.3}$ \\
\bottomrule
\end{tabular}
}
\label{Sec5_Table_JTVAE_GA_results}
\end{center}
\vspace{-4mm}
\end{table}
\section{Conclusion}

Black-box optimization in the latent space of VAEs can be substantially enhanced by leveraging the uncertainty of the decoder to avoid regions that will eventually lead to an invalid or low-quality decodings. This is even more critical when one has to explore regions that are far away from the VAE prior to uncover optimal objects. 
As a future research direction, leveraging the uncertainty of the property network jointly with constraints on the decoder uncertainty may make the search in latent even more robust to invalid decodings. 
Lastly, the importance sampling-based estimator introduced in this work is general-purpose and may be relevant to other applications that would benefit from reliable epistemic uncertainty estimates for complex high-dimensional data (e.g., active learning or anomaly detection).

\newpage
\small
\bibliographystyle{unsrtnat}
\bibliography{bibs/references}

\newpage

\appendix

\clearpage
\section*{Appendix}
\section{Analysis of variance of uncertainty estimators}
\label{Appendix_Uncertainty_estimator}

We demonstrate the lower variance of the Importance sampling-based estimator compared to the naive Monte Carlo estimator, focusing on the Character VAE for molecular generation setting described in \S~\ref{Sec5_CVAE} and Appendix~\ref{Appendix_D_Molecule_CVAE}.

\textbf{Setup.} We sample 1,000 points at random in latent space from an isotropic Gaussian with fixed standard deviation $\sigma$ (we repeat the experiment for different values of the standard deviation). We assess at these points the mutual information between outputs (generated molecule SMILES) and decoder parameters with the Importance sampling-based estimator (IS-MI) described in \S~\ref{Sec3_Estimator}, and the naive Monte Carlo (MC-MI) equivalent. The MC-MI estimator is obtained directly from equation \ref{Equation_MI_pre_IS} by sampling output sequences $y_s$ uniformly at random from  $\mathcal{S}$, the space of all possible SMILES strings. As per the setting described in Appendix~\ref{Appendix_D_Molecule_CVAE}, we limit the length of molecular sequences to 120 characters from a vocabulary comprised of 34 elements (plus the padding character). Consequently, $\mathcal{S}$ is finite and we can uniformly sample from it by independently sampling characters at each position uniformly at random from the vocabulary.
Both estimators are computed by sampling a fixed number of decoder parameters using Monte Carlo dropout \cite{gal2016dropout} (we used 100 model samples in all experiments).

\begin{figure}[ht]
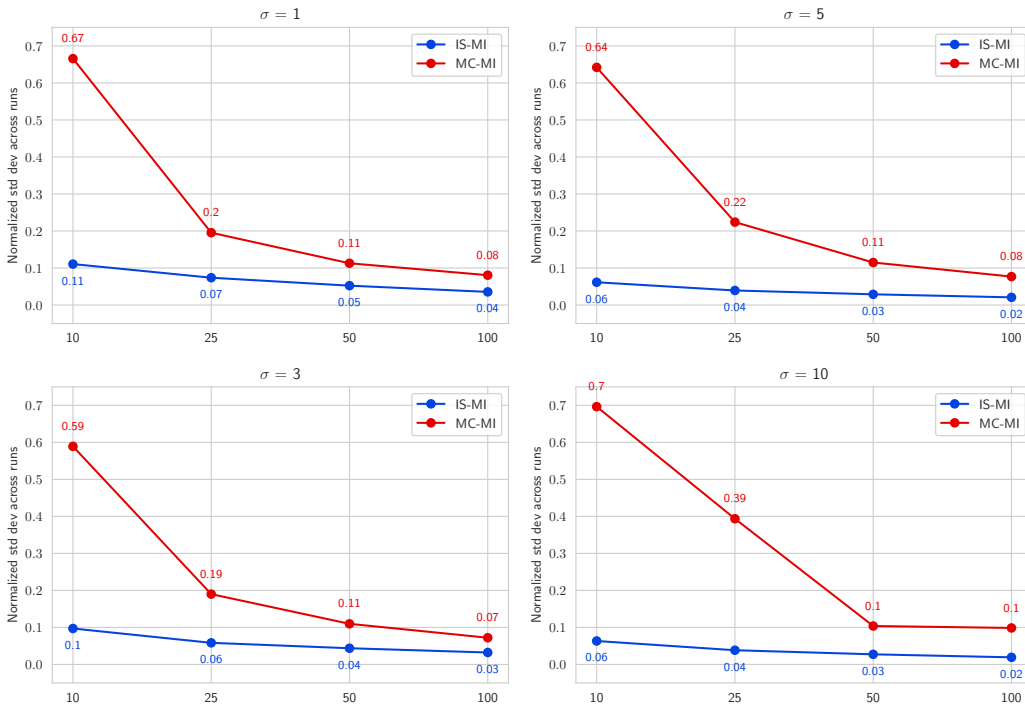

\vspace{0pt}
\resizebox{\textwidth}{!}{
    \includegraphics[scale=1]{Appendix/Images/Variance/CVAE_variance_std1.png}
    \includegraphics[scale=1]{Appendix/Images/Variance/CVAE_variance_std5.png}
}
\hfill
\resizebox{\textwidth}{!}{
    \includegraphics[scale=1]{Appendix/Images/Variance/CVAE_variance_std3.png}
    \includegraphics[scale=1]{Appendix/Images/Variance/CVAE_variance_std10.png}
}
\caption{\textbf{Variance analysis for uncertainty estimators} Comparison of the normalized standard deviations based on the number of sampled output sequences for the IS-MI Vs MC-MI estimators. We vary the standard deviation of the isotropic Gaussian used to sample points in latent space, from 1 (top left) to 10 (bottom right).}
\label{Appendix_Fig_variance_analysis_results}
\end{figure}

\textbf{Results.} We analyze the impact of the number of $y_s$ samples for each estimator on the variance of the corresponding estimators, measured over 10 independent runs. More specifically, we compute the standard deviation over the 10 runs, normalized by the estimator mean across runs. We observe that the IS-MI estimator has a normalized standard deviation 2-10x smaller than the MC-MI estimator across the different experiments (Fig.~\ref{Appendix_Fig_variance_analysis_results}).

\section{Digit generation experiments}
\label{Appendix_B_Digit_generation}

\subsection{Data}
\label{Appendix_B.1_Digit_generation_Data}

\textbf{Data source.} The MNIST \citep{Lecun1998MNIST} dataset consists of grayscale images (28x28 pixels) representing handwritten digits. The training dataset is comprised of 60k images and the test dataset is comprised of 10k images.

\textbf{Pre-processing.} We use the same train/test split as from the source listed in Table~\ref{Appendix_Data_sources}, and filter the data to keep images of 3 digits only. We then binarize the data, using a low pixel intensity threshold value ($10^{th}$ percentile of pixel intensity values observed on the training data). No data augmentation is used at train time nor at inference. We use the sum of all pixel intensities in the image as a proxy of the thickness of the digits, which provides strong empirical results.

\subsection{Model details}
\label{Appendix_B.2_Digit_generation_Model}

\begin{figure}[h]
    \centering
    \includegraphics[width=\textwidth]{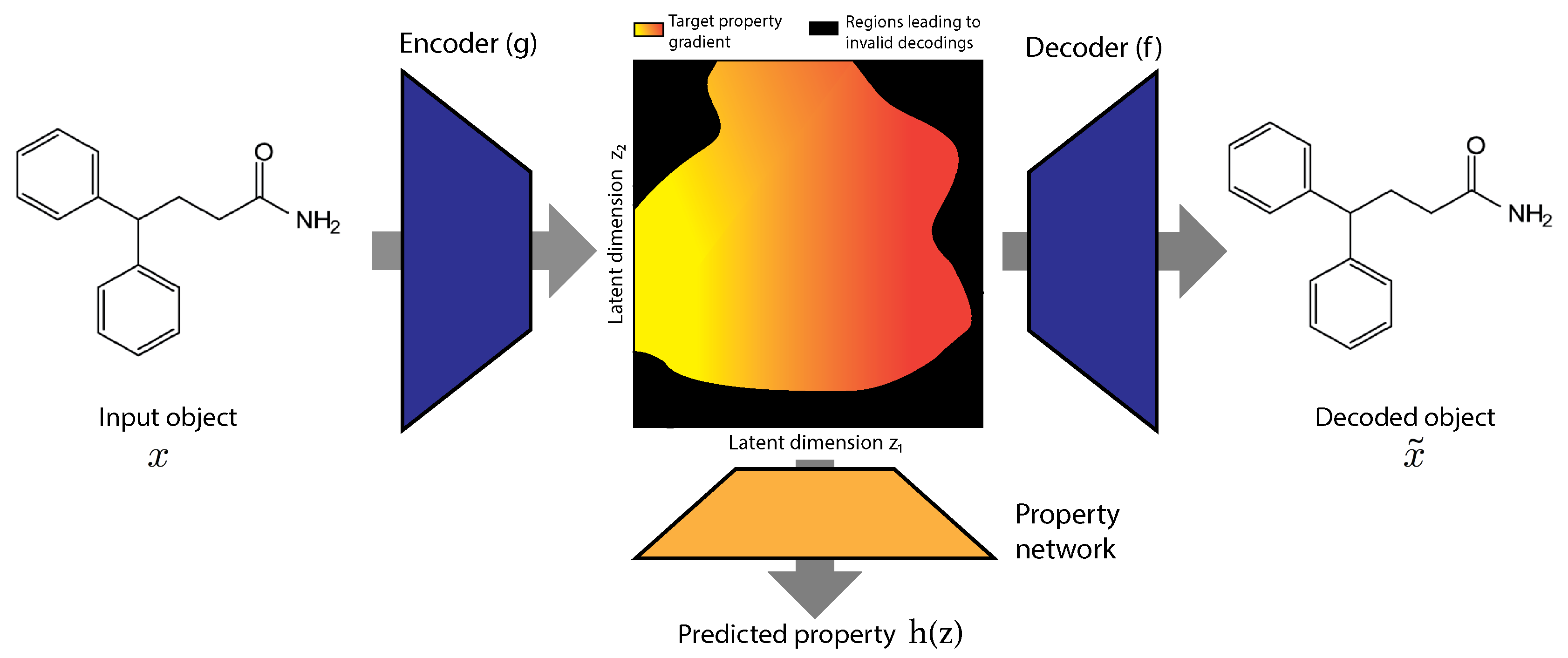}
    \caption{\textbf{Joint training architecture} A Variational autoencoder (VAE) is jointly trained with an auxiliary network predicting the value of the black-box objective from the latent space encoding. Optimization is then carried out in latent space via gradient ascent or Bayesian optimization}
    \label{Appendix_Figure_Joint_training_architecture}
\end{figure}

\textbf{Architecture.} We jointly train a variational autoencoder with an auxiliary network (the ``Property network``) predicting digit thickness based on latent representation (see  Fig.~\ref{Appendix_Figure_Joint_training_architecture}). For the VAE, we use a ``Conv-Deconv'' architecture\citep{Higgins2017}. All model parameters are summarized in table~\ref{Appendix_B_Table_model}.

\begin{table}[h]
\begin{center}
\caption{\textbf{Digit generation - Model architecture details}}
\resizebox{\textwidth}{!}{
\begin{tabular}{ll}
\toprule
\textbf{Component}  & \textbf{Description} \\
\toprule
\textbf{Encoder} & $\bullet$ 4 consecutive convolutional layers with 28, 56, 56 and 224 filters respectively, \\ 
& \quad kernel sizes 4,4,3 and 4 respectively, with batch norm and RELU activations \\
& \quad after each convolutional layer \\ 
& $\bullet$ Continuous latent space of dimension 2 \\
\midrule
\textbf{Decoder} &  $\bullet$ 4 2D-transposed convolutional layers with 224, 56, 56 and 28 filters respectively, \\
& \quad kernel sizes 4,3,4 and 4 respectively, with dropout 0.2 and RELU activations \\
& \quad after each convolutional layer \\ 
\midrule
\textbf{Property} &  $\bullet$ 3-layer feedforward network with 100 units each \\
\textbf{network} & $\bullet$ RELU activations (after each layer except the final one) and dropout 0.1 \\
\bottomrule
\end{tabular}
}
\label{Appendix_B_Table_model}
\end{center}
\end{table}

We assess the validity of generated digits with a separately trained Convolutional Neural Network (CNN) that learns to classify images of 3 digits (binary classification). This network is comprised of two convolution layers (with 23 and 64 filters respectively), followed by a max pool layer and two fully-connected layers (with 9,216 and 128 units respectively). RELU activations are used throughout, except for the final layer where a sigmoid activation is used for the binary classification.

\textbf{Training.} We train the joint architecture by minimizing the sum of the VAE ELBO and the mean squared error (MSE) on the thickness prediction task.
We use the Adam algorithm \citep{kingma2017adam} with a learning rate of $10^{-3}$, batch size of $512$, weight decay $10^{-5}$ for 300 epochs. We anneal the KL divergence with a sigmoid schedule for the first 30 epochs to avoid potential posterior collapse.

The independent CNN classifier network is trained by minimizing the binary cross entropy loss. We use the Adam optimizer, with learning rate $10^{-4}$, batch size 512 and train for 120 epochs.

\subsection{Uncertainty estimator}
\label{Appendix_B.3_Digit_generation_Uncertainty_estimator}

We compute the Importance sampling-based estimator following Algorithm~\ref{Sec3_Algorithm_IS-MI}, with $100$ samples from model parameters and $100$ binary images $y$ sampled from the importance distribution. We use Monte Carlo dropout \cite{gal2016dropout} to sample decoder parameters. 

\textbf{Uncertainty histograms. } We evaluate the IS-MI estimator values in different regions of latent space. We consider 4 distinct sets of points:
\vspace{-2mm}
\begin{itemize}
\setlength\itemsep{0em}
    \item \textbf{`Train':} Embeddings into latent space of 5k images sampled randomly from the train set;
    \item \textbf{`Test':} Embeddings into latent space of 5k images sampled randomly from the test set;
    \item \textbf{`Prior':} 5k random samples from the VAE prior;
    \item \textbf{`Far from prior':} 5k random samples from an isotropic Gaussian with standard deviation equal to 20.
\end{itemize}

\begin{figure}[ht]
\vspace{0pt}
\begin{center}
\resizebox{0.5\textwidth}{!}{
    \includegraphics[scale=1]{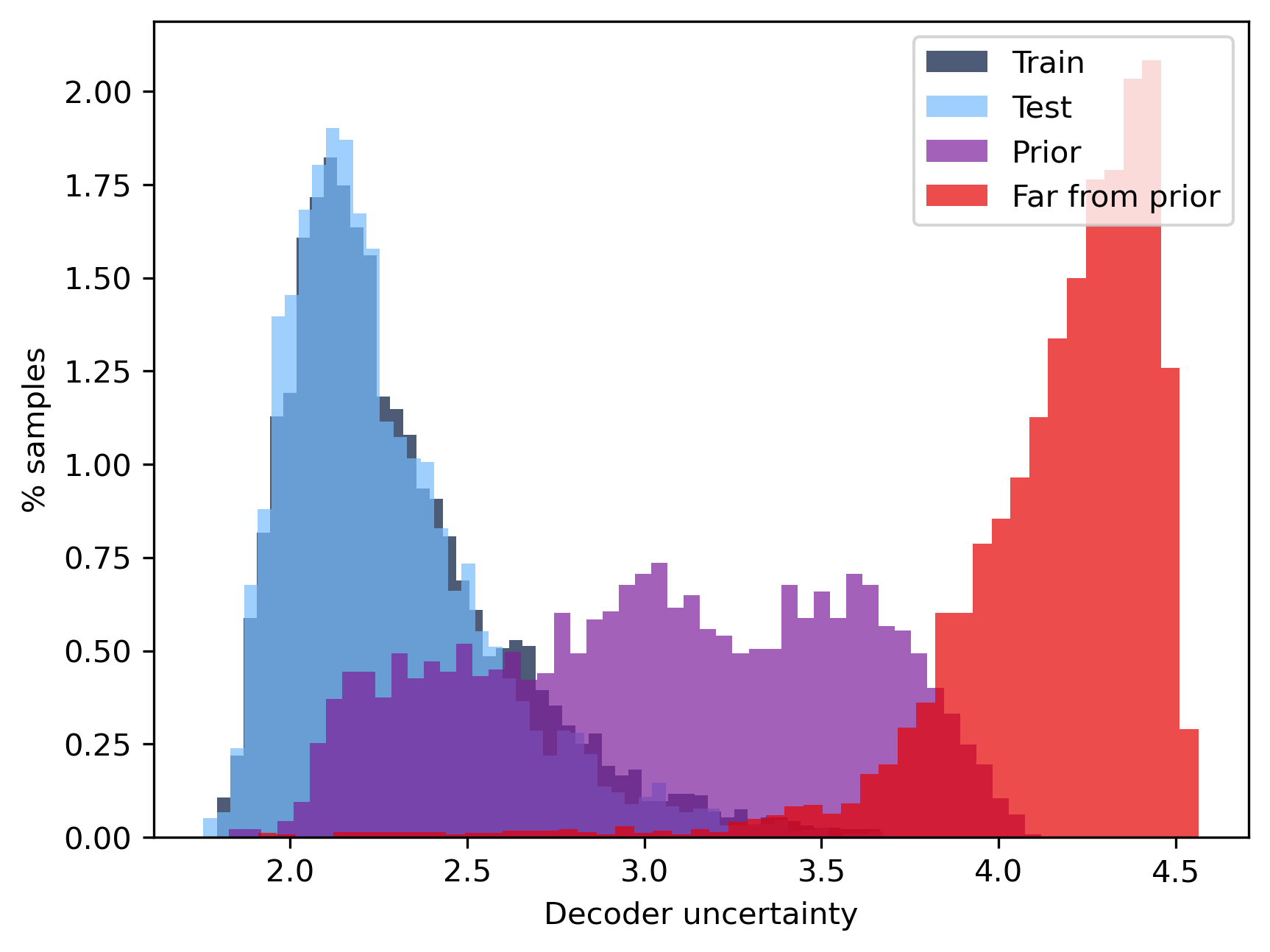}
}
\end{center}
\caption{\textbf{Digit generation - Decoder Uncertainty distribution} Distribution of decoder uncertainty values based on the IS-MI estimator, evaluated on the 4 distinct datasets defined in Appendix~\ref{Appendix_B.3_Digit_generation_Uncertainty_estimator}}
\label{Appendix_Fig_Digit_generation_histogram}
\end{figure}

We observe an almost perfect overlap between the distributions of decoder uncertainty values for the `Train' and `Test' sets, and both have a strong overlap with the `Prior' set (Fig.~\ref{Appendix_Fig_Digit_generation_histogram}). These first 3 sets are all fairly disjoint from the last set (`far from prior'), confirming the ability of the uncertainty estimators to properly identify `out-of-distribution' points.

\subsection{Optimization details}
\label{Appendix_B.4_Digit_generation_Optimization}

We follow the gradient ascent and Bayesian Optimization approaches described at the beginning of \S~\ref{Sec5_Experiments}. We provide below all hyperparameter values used for each method.

\textbf{Gradient ascent.} We start from 300 point sampled at random from the training set and then embedded in latent space. We perform 10 gradient update steps with a value of alpha equal to 50 (using notations from \S~\ref{Sec4_Uncertainty_guided_Optimization}). When censoring points based on the IS-MI estimator we used a conservative decoder uncertainty threshold, defined as the $75^{th}$ percentile of the IS-MI estimator values observed on training data.

\textbf{Bayesian optimization.} The single task Gaussian Process is initially trained on 1,000 images sampled at random from the training set that we then embed in latent space. We use the Expected Improvement as our acquisition function and sequentially generate 10 new digits overall (re-training the Gaussian Process after each acquisition). When leveraging the IS-MI estimator to guide the optimization, we censor proposal points based on the same uncertainty threshold as for gradient ascent ($75^{th}$ percentile of the IS-MI estimator values on training data). 

Results for the gradient ascent and Bayesian Optimization experiments are presented on Figure~\ref{Sec5_Fig_MNIST_Top_images}. Using the uncertainty of the decoder during gradient ascent of Bayesian Optimization helps maximizing thickness of the generated digits while preventing the them to be invalid white `blobs'.

\section{Arithmetic expression experiments}
\label{Appendix_C_Arithmetic_expression}

\subsection{Data}
\label{Appendix_C.1_Arithmetic_expression_Data}

\textbf{Data source.} We use the same dataset as in \citet{kusner2017grammar} which consists of univariate arithmetic expressions that are randomly generated from the following grammar:
\vspace{-4mm}
\begin{center}
\resizebox{0.6\textwidth}{!}{
\includegraphics[scale=1]{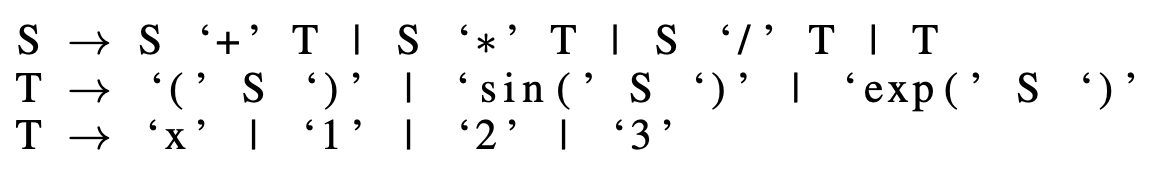}
}
\end{center}
\vspace{-4mm}
where S and T are non-terminals and the symbol `|' separates the possible production rules generated from each non-terminal. For instance, the following univariate arithmetic expressions can be generated from this grammar: $\sin(2)$, $x/(3+1)$, $2+x+\sin(1/2)$, and $x/2 * \exp(x)/\exp(2 * x)$. 

\textbf{Target.} The objective of the optimization in this setting is to find an expression that minimizes the mean squared error (MSE) with respect to a predefined target expression: $1/3 * x * \sin(x*x)$. Specifically, we measure the MSE between the target and proposal expressions over 1000 input  values $x$ that  are  linearly-spaced  between $-10$ and $10$.
Since the presence of exponentials in expressions may results in very large MSE values, the black-box objective to maximize is actually defined as $-\log(1+MSE)$.

\subsection{Model details}
\label{Appendix_C.2_Arithmetic_expression_Model}

\textbf{Architecture.} We jointly train a variational autoencoder and a property network which predicts the target defined above. For the VAE, we use an architecture identical to the CVAE in \citet{kusner2017grammar}, with a convolutional neural network (CNN) encoder and a Recurrent Neural Network (RNN) decoder. All model parameters are summarized in table~\ref{Appendix_C_Table_model}.

\begin{table}[h]
\begin{center}
\caption{\textbf{Arithmetic expressions approximation - Model architecture details}}
\resizebox{\textwidth}{!}{
\begin{tabular}{ll}
\toprule
\textbf{Component}  & \textbf{Description} \\
\toprule
\textbf{Encoder} & $\bullet$ 3 consecutive 1D convolutional layers with 2,3 and 4 filters respectively (kernel size 5),\\ 
& \quad with batch norm and RELU activations after each convolutional layer \\ 
& $\bullet$ Continuous latent space of dimension 25 \\
\midrule
\textbf{Decoder} &  $\bullet$ A stack of 3 Gated recurrent unit (GRU) layers \citep{cho2014properties} with hidden dimension 100, \\
& \quad dropout 0.2 (between layers) and RELU activations except for the last layer which \\ 
& \quad has a softmax activation over the arithmetic expressions vocabulary \\
\midrule
\textbf{Property} &  $\bullet$ 3-layer feedforward network with 200 units each \\
\textbf{network} & $\bullet$ RELU activations (after each layer except the final one) and dropout 0.2 \\
\bottomrule
\end{tabular}
}
\label{Appendix_C_Table_model}
\end{center}
\end{table}

\textbf{Training.} We train the joint architecture by minimizing the sum of the VAE ELBO and $\log(1+MSE)$ between an expression and the target one (as per Appendix~\ref{Appendix_C.1_Arithmetic_expression_Data}).
We minimize the loss with the Adam algorithm \citep{kingma2017adam} with a learning rate of $10^{-3}$ and batch size of $600$ for 80 epochs. We annealed the KL divergence with a sigmoid schedule for the first 10 epochs to avoid potential posterior collapse.

\subsection{Uncertainty estimators}
\label{Appendix_C.3_Arithmetic_expression_Uncertainty_estimator}

\begin{figure}[ht]
\vspace{0pt}
\resizebox{0.5\textwidth}{!}{
    \includegraphics[scale=1]{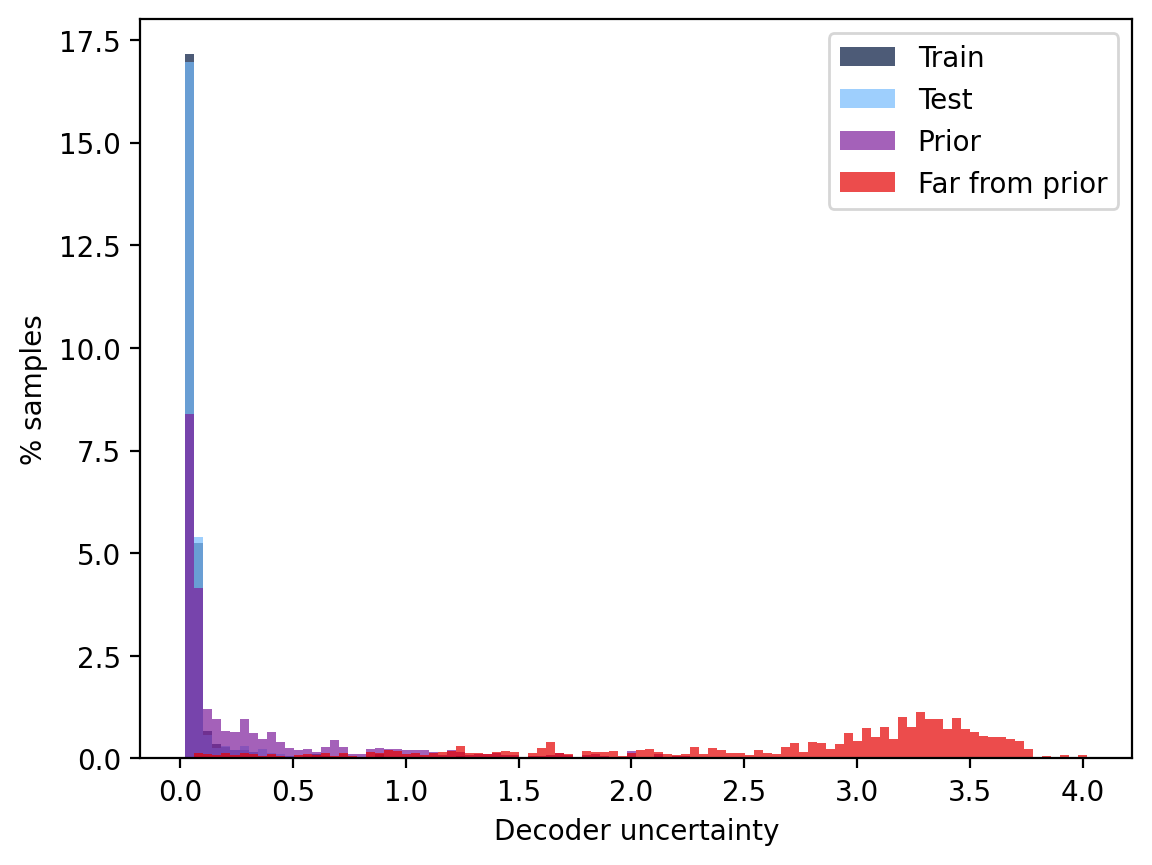}
}
\hfill
\resizebox{0.48\textwidth}{!}{
    \includegraphics[scale=1]{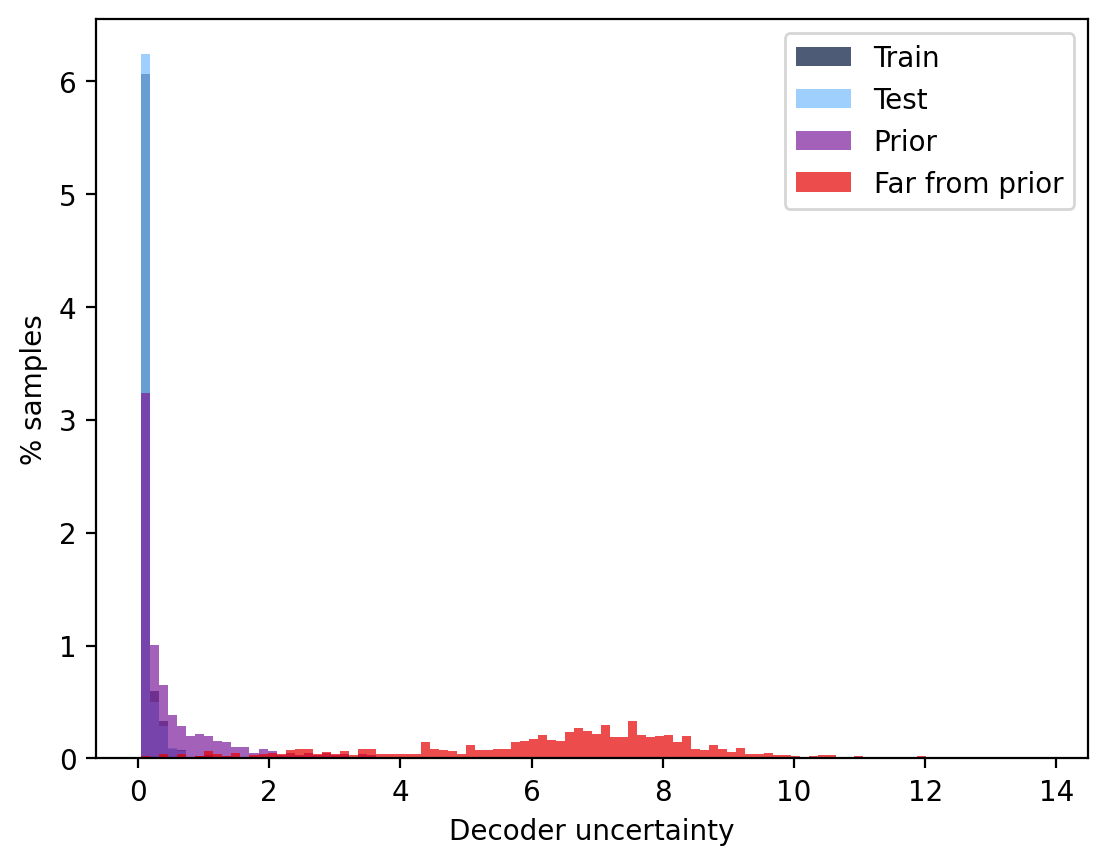}
}
\caption{\textbf{Arithmetic expressions approximation - Decoder Uncertainty distribution}. Analysis for the IS-MI estimator (left) and the TI-MI estimator (right). Both provide relatively good separation of `out-of-distribution' points in this setting.}
\label{Appendix_C_Fig_decoder_uncertainty_estimator}
\end{figure}

We replicate an analysis analogous to the one described in Appendix~\ref{Appendix_B.3_Digit_generation_Uncertainty_estimator}, sampling 1k points in latent space for each of the 4 datasets. The IS-MI estimator is computed with $100$ samples from model parameters via Monte Carlo dropout and $100$ expressions samples from the importance distribution. Similarly to what we observed before, the distribution of decoder uncertainty values for the `Train', `Test' and `Prior' sets have a strong overlap and are fairly disjoint from the `Far from prior' distribution (Fig.~\ref{Appendix_C_Fig_decoder_uncertainty_estimator}).

\subsection{Detailed optimization results}
\label{Appendix_C.4_Arithmetic_expression_Optimization}
We follow the gradient ascent and Bayesian Optimization approaches described at the beginning of \S~\ref{Sec5_Experiments}. We provide below all hyperparameter values used in each method.

\textbf{Gradient ascent.} We start from 500 point sampled at random from the training set and then embedded in latent space. We perform 10 gradient update steps with a value of alpha equal to 10 (using notations from \S~\ref{Sec4_Uncertainty_guided_Optimization}). When censoring points based on the IS-MI estimator we used a decoder uncertainty threshold defined as the $95^{th}$ percentile of the IS-MI estimator values observed on training data. Detailed results in Table~\ref{Appendix_C_Table_AE_GA_results}. 

\begin{table}[h]
\begin{center}
\caption{\textbf{Arithmetic expression - Gradient ascent results.}}
\begin{tabular}{cccccc}
\toprule
\textbf{Decoder} &  \multicolumn{4}{c}{\textbf{- log(1+MSE)}} & \textbf{Validity} \\
\textbf{uncertainty} & \textbf{Top 1} $\uparrow$ & \textbf{Top 2} $\uparrow$ & \textbf{Top 3} $\uparrow$ & \textbf{Avg. top 10} $\uparrow$ & \textbf{(\%)} $\uparrow$ \\
\toprule
None & $-0.32 \pm 0.04$& $-0.38 \pm 0.02$& $-0.42 \pm 0.01$& $-0.46 \pm 0.02$& $60\% \pm 0.6\%$\\
NLLP & $-0.22 \pm 0.05$& $-0.37 \pm 0.02$& $-0.39 \pm 0.01$& $-0.4 \pm 0.01$& $63\% \pm 0.8\%$\\
TI-MI & $-0.17 \pm 0.05$& $\textbf{-0.31} \pm \textbf{0.05}$& $-0.35 \pm 0.03$& $\textbf{-0.36} \pm \textbf{0.01}$& $\textbf{99\%} \pm \textbf{0.1\%}$\\
IS-MI & $\textbf{-0.13} \pm \textbf{0.04}$& $-0.32 \pm 0.03$& $\textbf{-0.33} \pm \textbf{0.04}$& $\textbf{-0.36} \pm \textbf{0.01}$& $98\% \pm 0.2\%$\\
\bottomrule
\end{tabular}
\label{Appendix_C_Table_AE_GA_results}
\end{center}
\end{table}

\textbf{Bayesian optimization.} The single task Gaussian Process is initially trained on 500 expressions sampled at random from the training set that we then embed in latent space. We sequentially generate 250 new arithmetic expressions (re-training the Gaussian Process after each acquisition). When using the IS-MI estimator to guide the optimization, we censor proposal points based on the same uncertainty threshold as for gradient ascent (ie., $95^{th}$ percentile of the IS-MI estimator values observed on training data). Detailed results in Table~\ref{Appendix_C_Table_AE_BO_results}.

\begin{table}[h]
\begin{center}
\caption{\textbf{Arithmetic expression - Bayesian Optimization results.}}
\begin{tabular}{cccccc}
\toprule
\textbf{Decoder} &  \multicolumn{4}{c}{\textbf{- log(1+MSE)}} & \textbf{Validity} \\
\textbf{uncertainty} & \textbf{Top 1} $\uparrow$ & \textbf{Top 2} $\uparrow$ & \textbf{Top 3} $\uparrow$ & \textbf{Avg. top 10} $\uparrow$ & \textbf{(\%)} $\uparrow$ \\
\toprule
None. & $-0.57 \pm 0.06$ & $-0.69 \pm 0.05$ & $-0.78 \pm 0.05$ & $-0.96 \pm 0.04$ & $77\% \pm 0.6\%$\\
NLLP & $-0.61 \pm 0.03$ & $-0.75 \pm 0.05$ & $-0.81 \pm 0.05$ & $-0.97 \pm 0.05$ & $76\% \pm 0.8\%$\\
TI-MI & $\textbf{-0.40} \pm \textbf{0.07}$ & $-0.55 \pm 0.03$ & $-0.64 \pm 0.05$ & $-0.72 \pm 0.06$ & $96\% \pm 0.5\%$\\
IS-MI & $-0.41 \pm 0.05$ & $\textbf{-0.52} \pm \textbf{0.05}$ & $\textbf{-0.59} \pm \textbf{0.05}$ & $\textbf{-0.70} \pm \textbf{0.05}$ & $\textbf{98\%} \pm \textbf{0.5\%}$\\
\bottomrule
\end{tabular}
\label{Appendix_C_Table_AE_BO_results}
\end{center}
\end{table}

In both the gradient ascent and Bayesian Optimization experiments, we obtain higher values of the black-box objective as well as a higher proportion of valid generated expressions (nearing 100\% in both cases) when leveraging decoder uncertainty. We get comparable results with the IS-MI and TI-MI estimators in this setting with relatively short sequences.


\section{Molecule generation experiments with CVAE}
\label{Appendix_D_Molecule_CVAE}

\subsection{Data}
\label{Appendix_D.1_Molecule_CVAE_Data}

\textbf{Data source.} We use a dataset of 250k drug-like molecules from the ZINC database \cite{Irwin2012ZINCAF}. Each molecule is represented via its SMILES representation \cite{weininger88smiles}, ie. as a sequence of characters (from a vocabulary of 34 elements, plus the padding character). Following \cite{Gomez_Bombarelli_2018}, molecule length is capped at 120, and shorter strings are space-padded to this length.

\textbf{Target.} The black-box objective in this set of experiments is the `penalized logP', defined as the octanol-water partition coefficient penalized by the synthetic accessibility score and the number of long cycles. We follow prior work \cite{kusner2017grammar, dai2018syntaxdirected, jin2019junction, liu2020chanceconstrained} and compute this metric as follows:
\begin{equation}
    \text{Penalized $\log$ P}(x) = \widehat{\log P(x)} - \widehat{SAS(x)} - \widehat{cycle(x)}
\end{equation}
where $\log P(x)$ is the octanol-water partition coefficient, $SAS(x)$ is the synthetic accessibility score, $cycle(x)$ counts the number of rings that have more than six atoms, and the\; $\widehat{    }$\; operator represents the standard normalization based on the raw training subset from ZINC (ie. subtracting the mean of the training set, and dividing by the standard deviation).

\subsection{Model details}
\label{Appendix_D.2_Molecule_CVAE_Model}

\textbf{Architecture.} We adopt a model architecture similar to \citet{Gomez_Bombarelli_2018}: the encoder is comprised of 3 convolutional layers, the decoder is composed of a stack of 3 GRU layers \citep{cho2014properties} and the property network has a simple feed forward architecture with 3 hidden layers. A detailed description is provided in Table~\ref{Appendix_D_Table_model}.

\begin{table}[h]
\begin{center}
\caption{\textbf{Molecular generation with CVAE - Model architecture details}}
\resizebox{\textwidth}{!}{
\begin{tabular}{ll}
\toprule
\textbf{Component}  & \textbf{Description} \\
\toprule
\textbf{Encoder} & $\bullet$ 3 consecutive 1D convolutional layers with 9,9 and 10 filters respectively,\\ 
& \quad kernel sizes 9,9 and 11 respectively, with batch norm and RELU activations \\
& \quad after each convolutional layer \\ 
& $\bullet$ Continuous latent space of dimension 56 \\
\midrule
\textbf{Decoder} &  $\bullet$ A stack of 3 Gated recurrent unit (GRU) layers \citep{cho2014properties} with hidden dimension 500, \\
& \quad with dropout 0.2 (between layers) and RELU activations except for the last \\ 
& \quad layer which has a softmax activation over the SMILES vocabulary.\\
& $\bullet$ At each step, the character generated at the previous step is concatenated \\ 
& \quad with the latent embedding and fed as input\\
\midrule
\textbf{Property} &  $\bullet$ 3-layer feedforward network with 1,000 units each \\
\textbf{network} & $\bullet$ RELU activations (after each layer except the final one) and dropout 0.2 \\
\bottomrule
\end{tabular}
}
\label{Appendix_D_Table_model}
\end{center}
\end{table}

\textbf{Training. } The total loss we minimize is the sum of the VAE ELBO and the MSE loss on the black-box property prediction task.
We train the network with the Adam algorithm \cite{kingma2017adam} with a learning rate of $5.10^{-4}$ (reduced by a factor 2 with a patience of 10 epochs) for 150 epochs total. We anneal the KL divergence with a sigmoid schedule for the first 30 epochs to avoid potential posterior collapse.
We also use teacher forcing on the character sampled at each time step in the decoder during training, and gradient clipping (upper bound set to 10) to avoid exploding gradients.

\subsection{Uncertainty estimators}
\label{Appendix_D.3_Molecule_CVAE_Uncertainty_estimator}
\begin{figure}[ht]
\vspace{0pt}
\resizebox{0.5\textwidth}{!}{
    \includegraphics[scale=1]{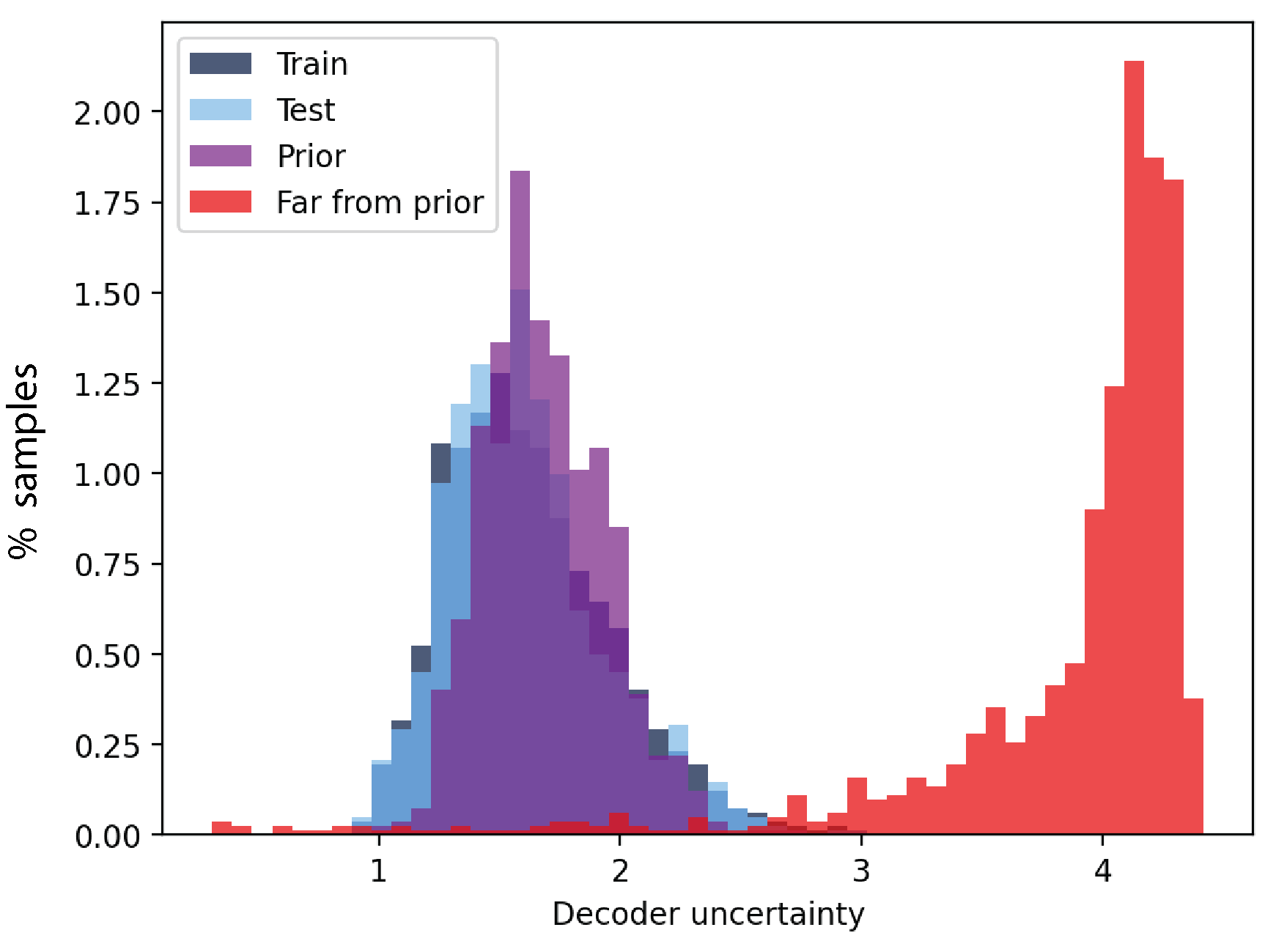}
}
\hfill
\resizebox{0.48\textwidth}{!}{
    \includegraphics[scale=1]{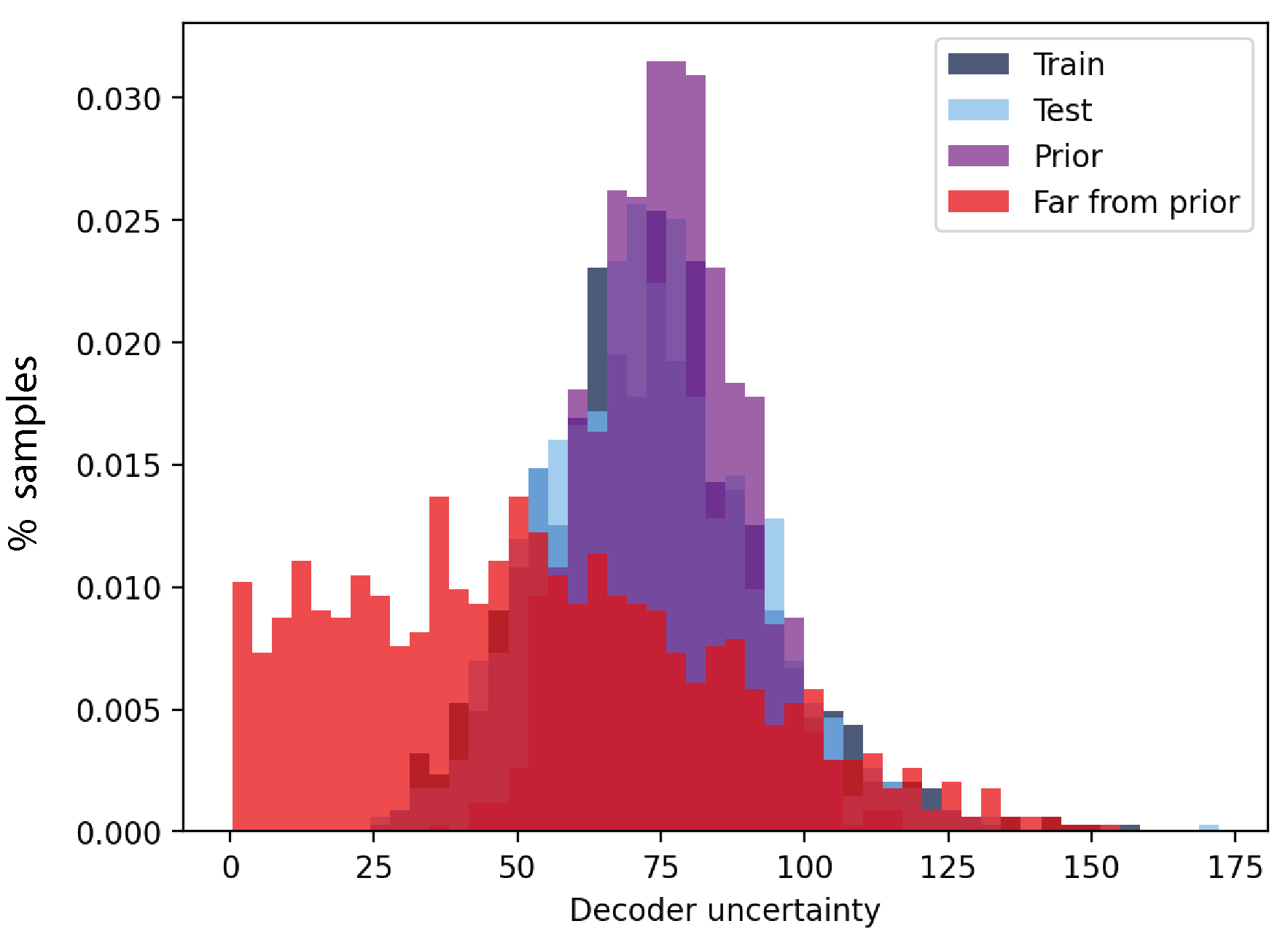}
}
\caption{\textbf{Molecular generation with CVAE - Decoder uncertainty distribution}. Analysis for the IS-MI estimator (left) and the TI-MI estimator (right)}
\label{Appendix_D_Fig_decoder_uncertainty_estimator}
\end{figure}

Similar to the previous two experimental settings, we first assess our uncertainty estimator by examining the distribution of its values on the 4 datasets defined in Appendix~\ref{Appendix_B.3_Digit_generation_Uncertainty_estimator}, using 1k samples for each dataset. We compute the IS-MI estimator based on Algorithm~\ref{Sec3_Algorithm_IS-MI}, use Monte Carlo dropout sampling to obtain 100 samples from decoder parameters, and sample 100 molecule SMILES from the importance distribution (Fig.~\ref{Appendix_D_Fig_decoder_uncertainty_estimator}). In this setting, the TI-MI estimator provides very poor uncertainty estimates as it assigns very low uncertainty values for a majority of `out-of-distribution' points that were sampled far from the prior (Fig.\ref{Appendix_D_Fig_decoder_uncertainty_estimator}).
This is consistent with what we see in Fig.\ref{Fig_CVAE_Uncertainty_estimator}b. In this experiment, we analyze the proportion of valid decodings when keeping the x\% points we are most certain about, based on the various estimators considered (ie. IS-MI, NLLP and TI-MI). If low uncertainty for a given estimator corresponds to high validity, then the \% of valid decodings should increase as we keep a narrower set of most confident points. This is what we observe for the IS-MI estimator, unlike the TI-MI estimator which incorrectly selects points leading to invalid decodings as the lowest uncertainty points.

\subsection{Detailed optimization results}
\label{Appendix_D.4_Molecule_CVAE_Optimization}
\textbf{Gradient ascent.} We start from 200 point sampled at random from the training set and embedded in latent space. We perform 10 gradient update steps with a value of alpha equal to 20. In the experiments where we impose a maximum threshold on the decoder uncertainty, we set that threshold as the $99^{th}$ percentile of corresponding estimator values observed on the training data. Leveraging the IS-MI estimator during optimization helps reaching higher values of the black-box objective (Table~\ref{Appendix_D_Table_CVAE_GA_results}). While constraints imposed with the NLLP baselines do help maintaining a higher \% of valid decodings, using this baseline is detrimental to optimization performance.

\begin{table}[h]
\begin{center}
\caption{\textbf{Molecular generation with CVAE - Gradient ascent results.}}
\begin{tabular}{cccccc}
\toprule
\textbf{Decoder} &  \multicolumn{4}{c}{\textbf{Penalized logP}} & \textbf{Validity} \\
\textbf{uncertainty} & \textbf{Top 1} $\uparrow$ & \textbf{Top 2} $\uparrow$ & \textbf{Top 3} $\uparrow$ & \textbf{Avg. top 10} $\uparrow$ & \textbf{(\%)} $\uparrow$ \\
\toprule
None & $5.1 \pm 0.2$& $4.6 \pm 0.2$& $4.1 \pm 0.1$& $1.6 \pm 0.5$& $4.1\% \pm 0.4\%$ \\
NLLP & $4.8 \pm 0.1$& $4.6 \pm 0$& $4.4 \pm 0$& $4.3 \pm 0$& $\textbf{56.6\%} \pm \textbf{0.9\%}$ \\
TI-MI & $3.3 \pm 2$& $4.6 \pm 0.1$& $4.4 \pm 0.1$& $0.8 \pm 1.8$& $4.1\% \pm 0.5\%$ \\
IS-MI & $\textbf{5.8} \pm \textbf{0.2}$& $\textbf{5.4} \pm \textbf{0.1}$& $\textbf{5.1} \pm \textbf{0.1}$& $\textbf{4.9} \pm \textbf{0.1}$& $21.5\% \pm 1\%$ \\
\bottomrule
\end{tabular}
\label{Appendix_D_Table_CVAE_GA_results}
\end{center}
\end{table}

\textbf{Bayesian Optimization.} We train a single task Gaussian Process (GP) on 500 points sampled at random from the training set and embedded in latent. We use the Expected Improvement as our acquisition function, sequentially generate 100 new molecules (and re-train the GP after each acquisition). Similar to our gradient ascent experiments, we set the uncertainty threshold to the $99^{th}$ percentile of corresponding estimator values observed on the training data. We observe that when using the IS-MI estimator, we not only increase the \% of valid decodings by 1.5-10x compared to baselines, but we also reach higher values of the `penalized logP' objective across all settings (Table.~\ref{Appendix_D_Table_CVAE_BO_results}). These results are robust to the choice of distribution decile used to define the uncertainty threshold, as shown in Table.\ref{Appendix_D_Table_CVAE_BO_threshold_analysis}.

\begin{table}[h]
\begin{center}
\caption{\textbf{Molecular generation with CVAE - Bayesian Optimization results.} NA values in the table corresponds to no valid molecule decoded across the 10 independent runs.}
\resizebox{\textwidth}{!}{
\begin{tabular}{ccccccc}
\toprule
\textbf{Search} & \textbf{Decoder} &  \multicolumn{4}{c}{\textbf{Penalized logP}} & \textbf{Validity} \\
\textbf{bounds} & \textbf{uncertainty} & \textbf{Top 1} $\uparrow$ & \textbf{Top 2} $\uparrow$ & \textbf{Top 3} $\uparrow$ & \textbf{Avg. top 10} $\uparrow$ & \textbf{(\%)} $\uparrow$ \\
\toprule
5 & None & $4.0 \pm 0.2$& $3.5 \pm 0.2$& $3.2 \pm 0.2$& $2.5 \pm 0.2$& $21.9\% \pm 1.4\%$ \\
& NLLP & $4.2 \pm 0.2$& $3.6 \pm 0.1$& $3.2 \pm 0.1$& $2.7 \pm 0.1$& $29.6\% \pm 1.3\%$ \\
& TI-MI & $4.1 \pm 0.2$& $3.5 \pm 0.2$& $3.1 \pm 0.1$& $2.3 \pm 0.1$& $21\% \pm 0.8\%$ \\
& IS-MI & $\textbf{4.5} \pm \textbf{0.2}$& $\textbf{3.7} \pm \textbf{0.2}$& $\textbf{3.5} \pm \textbf{0.2}$& $\textbf{3.0} \pm \textbf{0.1}$& $\textbf{33.2\%} \pm \textbf{1.8\%}$ \\
\midrule
10 & None & $3.9 \pm 1.1$& $-1.9 \pm 6.9$& $-14.4 \pm 4$& $-2.3 \pm 2.8$& $1.1\% \pm 0.4\%$ \\
& NLLP & $2.9 \pm 0.7$& $0.2 \pm 1.4$& $2.3 \pm 0.8$& $0.5 \pm 0.8$& $2.8\% \pm 0.7\%$ \\
& TI-MI & $5.9 \pm 3.3$& $-1.9 \pm 1.7$& $0.2 \pm 0.7$& $1.1 \pm 1.5$& $1.6\% \pm 0.4\%$ \\
& IS-MI & $\textbf{6.6} \pm \textbf{0.5}$& $\textbf{4.6} \pm \textbf{0.6}$& $\textbf{3.6} \pm \textbf{0.3}$& $\textbf{1.6} \pm \textbf{0.8}$& $\textbf{10.6}\% \pm \textbf{0.8\%}$ \\
\midrule
15 & None & $10.3 \pm 3.9$& $-3.0 \pm 2.7$& NA & $5.0 \pm 2.6$& $1\% \pm 0.3\%$ \\
& NLLP & $3.9 \pm 2.3$& $-4.6 \pm 4.9$& NA & $0.8 \pm 1.2$& $1\% \pm 0.3\%$ \\
& TI-MI & $6.7 \pm 3.6$& $0.0 \pm 1.7$& NA & $6.4 \pm 3.9$& $1.1\% \pm 0.3\%$ \\
& IS-MI & $\textbf{27.6} \pm \textbf{2.1}$& $\textbf{15.4} \pm \textbf{3.9}$& $\textbf{7.8} \pm \textbf{3.2}$& $\textbf{9.9} \pm \textbf{1.3}$& $\textbf{5.5\%} \pm \textbf{0.7\%}$ \\
\bottomrule
\end{tabular}
}
\label{Appendix_D_Table_CVAE_BO_results}
\end{center}
\end{table}

\begin{table}[h]
\begin{center}
\caption{\textbf{Molecular generation with CVAE - Impact of the choice of decoder uncertainty threshold on Bayesian Optimization results.} This analysis focuses on the IS-MI estimator only. NA values in the table corresponds to no valid molecule decoded. The second column represents the percentile of the IS-MI estimator values on the training data used to define the threshold for censored Bayesian Optimization.}
\resizebox{\textwidth}{!}{
\begin{tabular}{ccccccc}
\toprule
\textbf{Search} & \textbf{Uncertainty} &  \multicolumn{4}{c}{\textbf{Penalized logP}} & \textbf{Validity} \\
\textbf{bounds} & \textbf{threshold} & \textbf{Top 1} $\uparrow$ & \textbf{Top 2} $\uparrow$ & \textbf{Top 3} $\uparrow$ & \textbf{Avg. top 10} $\uparrow$ & \textbf{(\%)} $\uparrow$ \\
\toprule
5 & None & $4.0 \pm 0.2$& $3.5 \pm 0.2$& $3.2 \pm 0.2$& $2.5 \pm 0.2$& $21.9\% \pm 1.4\%$ \\
& Median & $4.2 \pm 0.1$& $\textbf{3.9} \pm \textbf{0.1}$& $\textbf{3.7} \pm \textbf{0.1}$& $\textbf{3.4} \pm \textbf{0.1}$& $\textbf{54.8}\% \pm \textbf{1.2\%}$ \\
& P90 & $4.3 \pm 0.1$& $3.8 \pm 0.1$& $3.6 \pm 0.1$& $3.3 \pm 0.1$& $48.2\% \pm 2.3\%$ \\
& P95 & $4.3 \pm 0.1$& $3.8 \pm 0.2$& $3.6 \pm 0.1$& $3.2 \pm 0.1$& $44.0\% \pm 1.7\%$ \\
& P99 & $\textbf{4.5} \pm \textbf{0.2}$& $3.7 \pm 0.2$& $3.5 \pm 0.2$& $3 \pm 0.1$& $33.2\% \pm 1.8\%$ \\
& Max & $4.3 \pm 0.2$& $3.6 \pm 0.2$& $3.1 \pm 0.2$& $2.6 \pm 0.2$& $26.9\% \pm 1.8\%$ \\
\midrule
10 & None & $3.9 \pm 1.2$& $-12.7 \pm 6.9$& $-14.4 \pm 4$& $-2.3 \pm 2.8$& $1.1\% \pm 0.4\%$ \\
& Median & $\textbf{7.6} \pm \textbf{0.7}$& $4.7 \pm 0.5$& $3.8 \pm 0.3$& $2.5 \pm 0.4$& $\textbf{12.4\%} \pm \textbf{0.7\%}$ \\
& P90 & $\textbf{7.6} \pm \textbf{0.7}$& $\textbf{4.8} \pm \textbf{0.5}$& $3.6 \pm 0.3$& $2.5 \pm 0.4$& $11.1\% \pm 0.7\%$ \\
& P95 & $7.0 \pm 0.9$& $4.5 \pm 0.6$& $\textbf{3.9} \pm \textbf{0.4}$& $\textbf{2.6} \pm \textbf{0.4}$& $12.0\% \pm 0.7\%$ \\
& P99 & $6.6 \pm 0.6$& $4.6 \pm 0.6$& $3.6 \pm 0.3$& $1.6 \pm 0.8$& $10.6\% \pm 0.8\%$ \\
& Max & $5.7 \pm 0.8$& $3.8 \pm 0.3$& $2.9 \pm 0.3$& $0.9 \pm 0.7$& $9.1\% \pm 0.8\%$ \\
\midrule
15 & None & $10.3 \pm 4.3$& $-3.0 \pm 2.7$& NA & $5.0 \pm 2.6$& $1.0\% \pm 0.3\%$ \\
& Median & $21.6 \pm 3.9$& $10.3 \pm 4.2$& $8.3 \pm 3.8$& $6.1 \pm 1.9$& $5.8\% \pm 0.8\%$ \\
& P90 & $24.0 \pm 3.3$& $14.1 \pm 4.2$& $7.8 \pm 3.4$& $7.6 \pm 1.7$& $5.8\% \pm 0.7\%$ \\
& P95 & $26.5 \pm 2.5$& $\textbf{17.2} \pm \textbf{4.1}$& $\textbf{8.9} \pm \textbf{4.1}$& $8.4 \pm 1.6$& $\textbf{6.2\%} \pm \textbf{0.7\%}$ \\
& P99 & $\textbf{27.6} \pm \textbf{2.2}$& $15.4 \pm 3.9$& $7.8 \pm 3.2$& $\textbf{9.9} \pm \textbf{1.3}$& $5.5\% \pm 0.7\%$ \\
& Max & $23.7 \pm 3.3$& $13.6 \pm 3.7$& $7.1 \pm 3.8$& $8.2 \pm 2.2$& $5.5\% \pm 0.7\%$ \\
\bottomrule
\end{tabular}
}
\label{Appendix_D_Table_CVAE_BO_threshold_analysis}
\end{center}
\end{table}
\vspace{-3mm}

\section{Molecule generation experiments with JTVAE}
\label{Appendix_E_Molecule_JTVAE}

\subsection{Data}
\label{Appendix_E.1_Molecule_JTVAE_Data}
We refer the reader to Appendix~\ref{Appendix_D.1_Molecule_CVAE_Data} as we used the same experimental setting as for the CVAE experiments, ie. we train our model on a subset of 250k drug-like molecules from the ZINC database, and seek to optimize the `penalized logP' metric.

\subsection{Model architecture}
\label{Appendix_E.2_Molecule_JTVAE_Model}

\textbf{Architecture.} We jointly train a Junction Tree VAE model (JT-VAE) with a property network predicting the `penalized logP' property based on latent representation, leveraging the same architecture design as in \citet{jin2019junction} for the constrained optimization task. The only difference we introduce is the incorporation of dropout layers in the junction tree decoder and graph decoder to allow sampling from the decoder parameters via Monte Carlo dropout (see Table~\ref{Appendix_E_Table_model}).

\begin{table}[h]
\begin{center}
\caption{\textbf{Molecular generation with JT-VAE - Model architecture details.} All components are identical to the JT-VAE architecture used for constrained optimization from \citet{jin2019junction}, except for the dropout layers detailed below}
\resizebox{\textwidth}{!}{
\begin{tabular}{ll}
\toprule
\textbf{Component}  & \textbf{Description} \\
\toprule
\textbf{Encoder} & $\bullet$ Junction tree encoder and molecular graph encoder  \\ 
& $\bullet$ Continuous latent space of dimension 56 \\
\midrule
\textbf{Decoder} &  $\bullet$ Junction tree decoder with dropout layer (0.2 drop rate) applied to the input and the output \\
& \quad of the GRU used for message passing, and dropout layer (0.2 rate) applied right before\\
& \quad  the final layer for both the topology prediction and node prediction networks \\ 
& $\bullet$ Molecular graph decoder with dropout layer (0.2 rate) applied right before the final layer \\
& \quad of the subgraph prediction network
\\
\midrule
\textbf{Property} &  $\bullet$ 2-layer feedforward network with 450 units and 1 unit resp., $\tanh$ activation for the first \\
\textbf{network} & \quad layer, no activation for the second, and dropout 0.2 before both layer\\
\bottomrule
\end{tabular}
}
\label{Appendix_E_Table_model}
\end{center}
\end{table}

\textbf{Training.} In line with the experiments discussed above, and following the same training procedure as per \citet{jin2019junction}, we minimize the sum of the VAE loss and the MSE on the black-box prediction task.

\subsection{Uncertainty estimators}
\label{Appendix_E.3_Molecule_JTVAE_Uncertainty_estimator}
We compute the IS-MI estimator based on Algorithm~\ref{Sec3_Algorithm_IS-MI}. Following notations from Algorithm~\ref{Sec3_Algorithm_IS-MI}, we sample a molecule $\tilde{y}_s$ by successively decoding from the junction tree decoder and then the graph decoder. We then decompose $\log{p_{s,m}}$ as the sum of the log probabilities corresponding to each prediction made by the junction tree decoder and graph decoder (for the graph outcome generated in the previous step), namely the topology and node predictions in the junction tree decoder, and the subgraph prediction in the graph decoder. 
Similar to the setting discussed in Appendix~\ref{Appendix_B.3_Digit_generation_Uncertainty_estimator}, we inspect the distribution of the IS-MI values obtained on the same 4 datasets (using 1k samples for each dataset), and estimate the mutual information with $100$ samples from decoder parameters and $100$ molecules sampled from the importance sampling distribution. We observe similar results as before: overlap between IS-MI distributions on the `Train', `Test' and `Prior' sets. This overlap is stronger between the first two sets as the embedding of the training data in latent does not necessarily follow a standard normal distribution after model training. The distribution of IS-MI values on the `Far from prior' set is disjoint from the first 3 sets, with the highest values obtained on this set.

\begin{figure}[h!]
    \centering
    \includegraphics[width=0.45\textwidth]{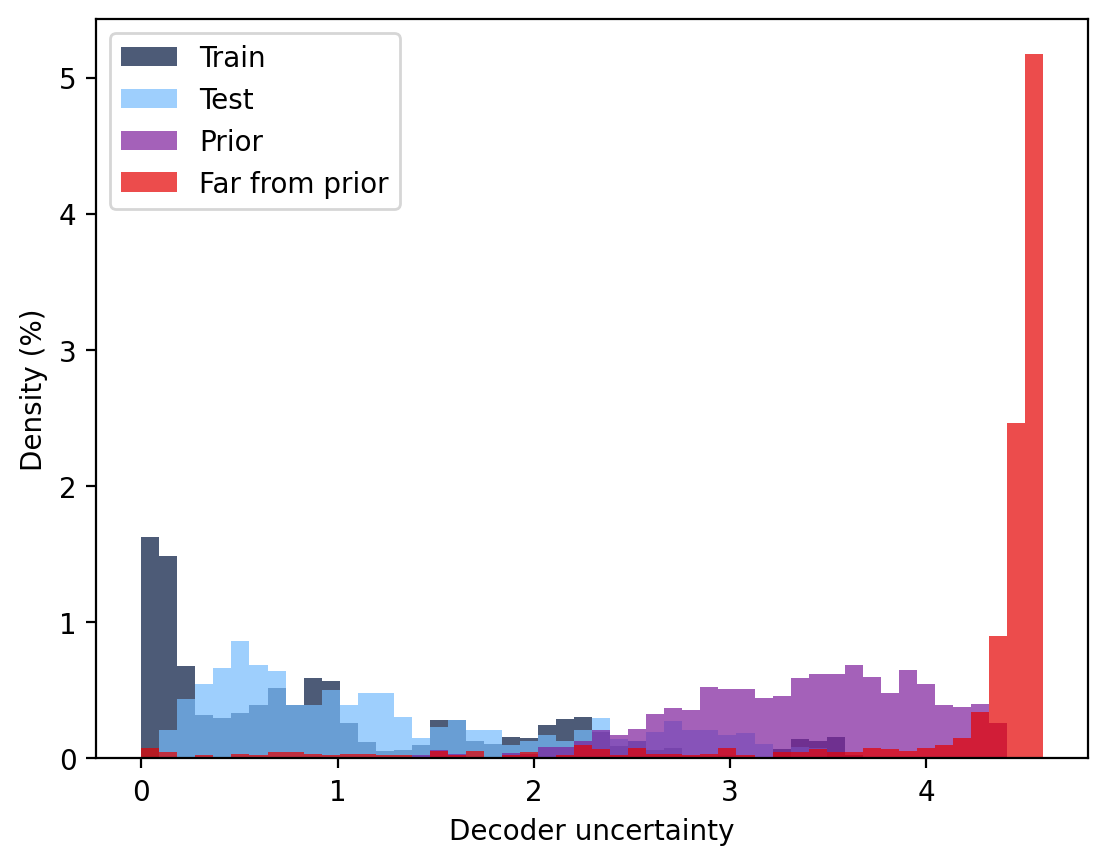}
    \caption{\textbf{Molecular generation with JT-VAE - Decoder uncertainty distribution.}}
    \label{Appendix_Fig_JTVAE_Uncertainty_estimator}
\end{figure}

\subsection{Detailed optimization results}

All molecules generated by the JT-VAE are valid by design. However, not all generated molecules will be of high \emph{quality}, as measured for example by the quality filters from \citet{Brown_2019} discussed in \S~\ref{Sec5_JTVAE}. Since in this molecular generation setting, our objective is to generate new molecules with high penalized logP values that could be used as potential drugs, we want to ensure that the candidate drugs we shortlist for further investigation also pass these quality filters. In each optimization experiment described below, we prioritize a small number of candidate molecules (eg., top 10 molecules with highest logP values) and use the quality filters from \citet{Brown_2019} as a proxy for the subsequent (costly) verification of these candidates by medicinal chemists. If a generated molecule does not pass these quality filters, its penalized logP value is assigned to a default value (eg., average penalized logP value on the training set). In all optimization experiments, we estimate mutual information with $100$ samples from decoder parameters, and a single molecule sampled from the importance distribution.

\begin{figure*}[h]
    \centering
    \includegraphics[width=5.5in,scale=1.0]{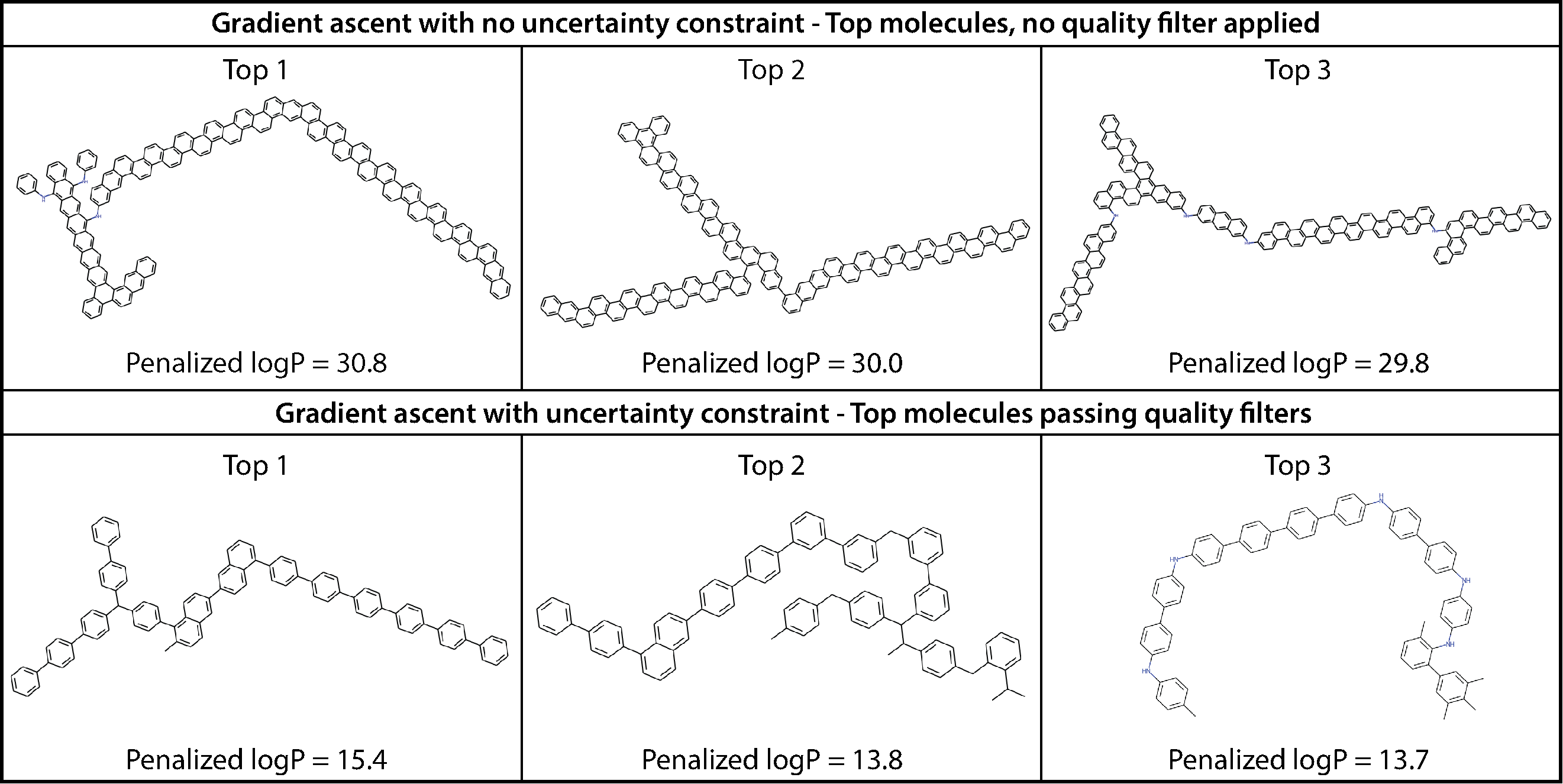}
    \caption{\textbf{Top molecules generated via gradient ascent with a JT-VAE} ($\alpha$ = 200).}
    \label{Fig_Appendix_JTVAE_GA_Top_molecules_generated}
\end{figure*}

\label{Appendix_E.4_Molecule_JTVAE_Optimization}
\textbf{Gradient ascent.} We start from 100 molecules sampled at random from the training set, that we then embed in latent space. We perform 100 gradient update steps with a large value of $\alpha$ (as per notations in \S~\ref{Sec4_Uncertainty_guided_Optimization}), eg., 100 or 200. This leads state-of-the-art performance in terms of penalized logP values (see Table~\ref{Appendix_E_Table_Molecular_generation_top_performance} and Fig.~\ref{Fig_Appendix_JTVAE_GA_Top_molecules_generated}). However, as we move `further away' in latent space, the quality of generated molecules tends to degrade.
By setting an upper bound on the uncertainty of the decoder (eg., $95^{th}$ percentile of IS-MI values observed on the training data) during optimization, we are able to generate molecules with both high penalized logP values and high quality (see Table~\ref{Appendix_E_Table_JTVAE_GA_results}). Selecting different uncertainty threshold values enable to reach different trade-offs between quality and black-box objective.
A similar approach with upper bounds in terms of NLLP values (eg., threshold defined as $95^{th}$ or $99^{th}$ percentiles of NLLP values on the training data) does help promoting high quality molecules but leads to much lower penalized logP values.

\begin{table}[h]
\begin{center}
\caption{\textbf{Molecular generation - Top optimization performance.} We achieve state-of-the-art performance on the molecular generation task using a JT-VAE model jointly trained with an auxiliary network predicting penalized logP from latent embeddings (as per \S 3.3 of \cite{jin2019junction}) and performing gradient ascent as described in \S~\ref{Sec4_Uncertainty_guided_Optimization}. Results were obtained by embedding in latent space 100 points selected at random from the test set and then performing 100 gradient updates with $\alpha = 200$. We report mean performance over 10 runs, as well as the best generated molecules across these 10 runs.}
\resizebox{\textwidth}{!}{
\begin{tabular}{llcccc}
\toprule
\textbf{Model} & \textbf{Optimization method} &  \multicolumn{3}{c}{\textbf{Penalized logP}} \\
&  & \textbf{Top 1} $\uparrow$ & \textbf{Top 2} $\uparrow$ & \textbf{Top 3} $\uparrow$\\
\toprule
JT-VAE \cite{jin2019junction} & Bayesian Optimization & 5.30 & 4.93 & 4.49 \\
MolDQN \cite{Zhou_2019MolDQN} & Reinforcement learning & 11.84 & 11.84 & 11.82 \\
GraphAF \cite{shi2020graphaf} & Reinforcement learning & 12.23 & 11.29 & 11.05 \\
CCGF \cite{liu2020chanceconstrained} & Chance-constrained optimization & 12.32 & 11.79 & 11.61 \\
ChemBO \cite{korovina2019chembo} & Bayesian Optimization & 18.39 & - & - \\
JT-VAE \cite{tripp2020sampleefficient} & Bayesian Optim. \& retraining (median of 5 runs) & 21.20 & 15.34 & 15.34 \\
JT-VAE \cite{tripp2020sampleefficient} & Bayesian Optim. \& retraining (best over 5 runs) & 27.84 & 27.59 & 27.21\\ 
\midrule
JT-VAE (ours) & Gradient ascent (mean of 10 runs) & 23.65	& 21.17	& 19.45\\
JT-VAE (ours) & Gradient ascent (best over 10 runs) & \textbf{30.81} & \textbf{30.00} & \textbf{29.82}\\
\bottomrule
\end{tabular}
\label{Appendix_E_Table_Molecular_generation_top_performance}
}
\end{center}
\end{table}
\begin{table}[h]
\begin{center}
\caption{\textbf{Molecular generation with JT-VAE - Gradient ascent results.} The `Uncertainty threshold' column represents the percentile of training values used to define the uncertainty threshold.}
\resizebox{\textwidth}{!}{
\begin{tabular}{ccccccccccc}
\toprule
$\alpha$ = 100\\
\toprule
\textbf{Decoder} & \textbf{Uncertainty} & \multicolumn{4}{c}{\textbf{Penalized logP - Before filters}} & \textbf{Quality} & \multicolumn{4}{c}{\textbf{Penalized logP - Passing filters}} \\
\textbf{uncertainty} & \textbf{threshold} & \textbf{Top 1} $\uparrow$ & \textbf{Top 2} $\uparrow$ & \textbf{Top 3} $\uparrow$ & \textbf{Avg. top 10} $\uparrow$ & \textbf{top 10 (\%)} $\uparrow$ & \textbf{Top 1} $\uparrow$ & \textbf{Top 2} $\uparrow$ & \textbf{Top 3} $\uparrow$ & \textbf{Avg. top 10} $\uparrow$ \\
\toprule
None & None & $\textbf{22.4} \pm \textbf{0.9}$ & $\textbf{19.9} \pm \textbf{0.6}$ & $\textbf{18.7} \pm \textbf{0.4}$ & $\textbf{16.6} \pm \textbf{0.3}$ & $3\% \pm 2\% $ & $4.3 \pm 2.2$ & $0.0 \pm 0.0$ & $0.0 \pm 0.0$ & $0.4 \pm 0.2$ \\
\midrule
NLLP & P95 & $3.4 \pm 0.1$ & $2.9 \pm 0.1$ & $2.6 \pm 0.1$ & $2.4 \pm 0.1$ & $71\% \pm 4\% $ & $3.3 \pm 0.2$ & $2.7 \pm 0.1$ & $2.5 \pm 0.1$ & $1.8 \pm 0.1$ \\
 & P99 & $3.8 \pm 0.1$ & $3.4 \pm 0.1$ & $3.2 \pm 0.1$ & $3.0 \pm 0.1$ & $\textbf{89\%} \pm \textbf{3\%} $ & $3.8 \pm 0.1$ & $3.3 \pm 0.1$ & $3.2 \pm 0.1$ & $2.7 \pm 0.1$ \\
 & Max & $4.5 \pm 0.1$ & $4.2 \pm 0.1$ & $4.0 \pm 0.1$ & $3.8 \pm 0.1$ & $82\% \pm 4\% $ & $4.4 \pm 0.1$ & $4.0 \pm 0.1$ & $3.8 \pm 0.1$ & $3.1 \pm 0.1$ \\
\midrule
IS-MI & P95 & $11.4 \pm 1.4$ & $8.1 \pm 0.3$ & $7.7 \pm 0.2$ & $7.6 \pm 0.2$ & $81\% \pm 5\% $ & $8.3 \pm 0.3$ & $7.7 \pm 0.2$ & $\textbf{7.4} \pm \textbf{0.2}$ & $\textbf{5.8} \pm \textbf{0.3}$ \\
 & P99 & $17.5 \pm 1.4$ & $13.3 \pm 0.9$ & $11.4 \pm 0.7$ & $10.3 \pm 0.5$ & $48\% \pm 6\% $ & $\textbf{9.9} \pm \textbf{0.7}$ & $\textbf{8.4} \pm \textbf{0.2}$ & $6.9 \pm 0.8$ & $3.9 \pm 0.4$ \\
 & Max & $19.4 \pm 1.1$ & $16.7 \pm 1.0$ & $15.1 \pm 0.9$ & $13.0 \pm 0.6$ & $19\% \pm 4\% $ & $\textbf{9.9} \pm \textbf{1.1}$ & $4.9 \pm 1.5$ & $2.6 \pm 1.2$ & $1.8 \pm 0.3$ \\
\toprule
$\alpha$ = 200\\
\toprule
\textbf{Decoder} & \textbf{Uncertainty} & \multicolumn{4}{c}{\textbf{Penalized logP - Before filters}} & \textbf{Quality} & \multicolumn{4}{c}{\textbf{Penalized logP - Passing filters}} \\
\textbf{uncertainty} & \textbf{threshold} & \textbf{Top 1} $\uparrow$ & \textbf{Top 2} $\uparrow$ & \textbf{Top 3} $\uparrow$ & \textbf{Avg. top 10} $\uparrow$ & \textbf{top 10 (\%)} $\uparrow$ & \textbf{Top 1} $\uparrow$ & \textbf{Top 2} $\uparrow$ & \textbf{Top 3} $\uparrow$ & \textbf{Avg. top 10} $\uparrow$ \\
\toprule
None & None & $\textbf{23.7} \pm \textbf{1.3}$ & $\textbf{21.2} \pm \textbf{0.8}$ & $\textbf{19.5} \pm \textbf{0.8}$ & $\textbf{17.0} \pm \textbf{0.6}$ & $1\% \pm 1\% $ & $1.2 \pm 1.2$ & $0.0 \pm 0.0$ & $0.0 \pm 0.0$ & $0.1 \pm 0.1$ \\
\midrule
NLLP & P95 & $3.0 \pm 0.1$ & $2.8 \pm 0.1$ & $2.7 \pm 0.1$ & $2.5 \pm 0.05$ & $82\% \pm 6\% $ & $3.0 \pm 0.1$ & $2.7 \pm 0.1$ & $2.6 \pm 0.1$ & $2.0 \pm 0.2$ \\
 & P99 & $3.3 \pm 0.1$ & $3.0 \pm 0.1$ & $2.8 \pm 0.1$ & $2.6 \pm 0.1$ & $80\% \pm 4\% $ & $3.2 \pm 0.1$ & $2.9 \pm 0.1$ & $2.7 \pm 0.1$ & $2.1 \pm 0.1$ \\
 & Max & $3.6 \pm 0.1$ & $3.2 \pm 0.1$ & $3.0 \pm 0.1$ & $2.7 \pm 0.1$ & $81\% \pm 4\% $ & $3.6 \pm 0.1$ & $3.1 \pm 0.1$ & $3.0 \pm 0.1$ & $2.2 \pm 0.1$ \\
\midrule
IS-MI & P95 & $8.4 \pm 0.8$ & $6.8 \pm 0.4$ & $6.4 \pm 0.4$ & $6.0 \pm 0.3$ & $\textbf{89\%} \pm \textbf{3\%} $ & $7.7 \pm 0.7$ & $6.5 \pm 0.3$ & $6.1 \pm 0.3$ & $\textbf{5.3} \pm \textbf{0.3}$ \\
 & P99 & $15.3 \pm 1.6$ & $9.5 \pm 1.0$ & $7.2 \pm 0.3$ & $7.1 \pm 0.4$ & $69\% \pm 4\% $ & $7.6 \pm 0.3$ & $\textbf{6.8} \pm \textbf{0.3}$ & $\textbf{6.3} \pm \textbf{0.3}$ & $4.0 \pm 0.1$ \\
 & Max & $20.5 \pm 1.3$ & $17.3 \pm 1.4$ & $13.8 \pm 1.2$ & $11.4 \pm 0.8$ & $31\% \pm 7\% $ & $\textbf{8.6} \pm \textbf{1.3}$ & $6.1 \pm 1.1$ & $3.4 \pm 1.1$ & $2.3 \pm 0.5$ \\
\bottomrule
\end{tabular}
}
\label{Appendix_E_Table_JTVAE_GA_results}
\end{center}
\end{table}

\textbf{Bayesian Optimization.} We train a single task Gaussian Process (GP) on 500 points sampled at random from the training set and embedded in latent. We use the Expected Improvement as our acquisition function, sequentially generate 500 new molecules (and re-train the GP after each acquisition). In experiments in which we impose an upper bound on decoder uncertainty, we set that bound as the $99^{th}$ percentile of decoder uncertainty values observed on the training data. Similar to what we observe in the gradient ascent experiments, leveraging the IS-MI estimator helps generating candidate molecules with both high `penalized logP' values and high quality (Table~\ref{Appendix_E_Table_JTVAE_BO_results}). The NLLP of proposal points at each step of the batch Bayesian Optimization tend to always be above the NLLP threshold ($99^{th}$ percentile of values on the training data). In these situations, we select the point with lowest NLLP value from the proposal batch as described in \S~\ref{Sec5_Experiments}. This explains why the results obtained with NLLP are closer to what we obtain without any constraint during optimization, and why we lose the ability to increase the quality of generated molecules in this setting.

\begin{table}[h]
\begin{center}
\caption{\textbf{Molecular generation with JT-VAE - Bayesian Optimization results.}}
\resizebox{\textwidth}{!}{
\begin{tabular}{cccccccccc}
\toprule
\textbf{Decoder} & \multicolumn{4}{c}{\textbf{Penalized logP - Before filters}} & \textbf{Quality} & \multicolumn{4}{c}{\textbf{Penalized logP - Passing filters}} \\
\textbf{uncertainty} & \textbf{Top 1} $\uparrow$ & \textbf{Top 2} $\uparrow$ & \textbf{Top 3} $\uparrow$ & \textbf{Avg. top 10} $\uparrow$ & \textbf{top 10 (\%)} $\uparrow$ & \textbf{Top 1} $\uparrow$ & \textbf{Top 2} $\uparrow$ & \textbf{Top 3} $\uparrow$ & \textbf{Avg. top 10} $\uparrow$ \\
\toprule
None & $\textbf{18.6} \pm \textbf{1.5}$ & $\textbf{14.1} \pm \textbf{1.8}$ & $\textbf{11.5} \pm \textbf{1.2}$ & $\textbf{9.6} \pm \textbf{0.7}$ & $18\% \pm 6\% $ & $7.1 \pm 2.2$ & $3.0 \pm 1.1$ & $2.2 \pm 1.1$ & $1.4 \pm 0.4$ \\
NLLP & $15.7 \pm 1.5$ & $12.2 \pm 0.8$ & $9.7 \pm 0.5$ & $8.7 \pm 0.4$ & $23\% \pm 5\% $ & $7.8 \pm 0.6$ & $4.6 \pm 1.0$ & $1.1 \pm 0.7$ & $1.5 \pm 0.3$ \\
IS-MI & $14.3 \pm 1.7$ & $9.2 \pm 1.0$ & $7.6 \pm 0.9$ & $6.3 \pm 0.4$ & $\textbf{47\%} \pm \textbf{3\%} $ & $\textbf{9.7} \pm \textbf{2.1}$ & $\textbf{4.8} \pm \textbf{0.4}$ & $\textbf{4.2} \pm \textbf{0.3}$ & $\textbf{2.1} \pm \textbf{0.3}$ \\
\bottomrule
\end{tabular}
}
\label{Appendix_E_Table_JTVAE_BO_results}
\end{center}
\end{table}

\vspace{-2mm}
\section{Discussion}

\vspace{-1mm}
\textbf{Strengths of the approach. } Leveraging the uncertainty of the decoder is a \emph{simple} yet \emph{effective} approach to promote the validity or quality of objects generated while optimizing a given black-box property in VAE latent space, without compromising on the final objective. It is simple because it does not impose changes to the model architecture nor the training procedure to work. The only requirement is the ability to sample from decoder parameters -- which we have achieved in this work via Monte Carlo dropout given its practical simplicity.
As we have demonstrated, the value of the approach is not limited to a particular decoder architecture, and the proposed scheme delivered compelling results across convolutional, recurrent and graph neural network decoders.
In several of the experimental settings, using the decoder uncertainty also helped attaining substantially higher values of the black-box objective during optimization (\S~\ref{Sec5_CVAE} in particular), as the optimization procedure could use the decoder uncertainty to avoid exploring regions which are doomed to lead to invalid decodings.

\textbf{Limitations. } The main limitation of the approach is the computation overhead resulting from the estimation of the decoder uncertainty. While the computation of the IS-MI estimator is easily parallelizable, it nonetheless increases the time required to complete each optimization step (this overhead depends on a number of factors including number of samples for IS-MI evaluation, decoder architecture and hardware used). 
In the setting we have considered, the black-box objective may be very expensive to evaluate (eg., a costly wet lab experiment, a time-consuming simulation). Consequently, the ability to generate strong candidates that satisfy non-obvious validity or quality requirements takes precedence over the computation time needed to generate them. 
Furthermore, the computation overhead may also not be a primary concern in situations where the uncertainty of the decoder helps achieve a superior optimum compared to what could have been achieved without it (eg., \S~\ref{Sec5_CVAE}). 
However, in settings in which the decoder uncertainty does not help reaching superior optima and in which the cost of evaluating candidate objects is low, then the decision of whether to leverage the decoder uncertainty or not should be based on the relative trade-off between validity gains during optimization and computation overhead.

\textbf{What we thought would work, but did not work. } In \S~\ref{Sec4_Uncertainty_guided_Optimization} we presented two possible approaches for incorporating the decoder uncertainty within Bayesian Optimization in latent space: the uncertainty censoring approach and the uncertainty-aware surrogate model approach. While we have used the former extensively across experiments in \S~\ref{Sec5_Experiments}, the latter has not delivered strong results. A possible interpretation would be that the latter approach may heavily penalize optimization directions for which the gradient of the black-box objective is aligned with the gradient of decoder uncertainty, while the former would allow moving along these directions up until the uncertainty threshold is exceeded.

\textbf{Societal impact. } We introduce a general approach to improve the black-box optimization of complex high-dimensional discrete objects in VAE latent space. The potential applications of this field are very broad -- from more effective drug or protein design to improved automatic program synthesis. The estimator of decoder uncertainty we have introduced may also be leveraged in a wide range of other areas that would benefit from reliable uncertainty estimates in high dimensional settings (eg., active learning, outlier detection).

\vspace{-2mm}
\section{Reproducibility}
\vspace{-1mm}
\textbf{Code and dependencies. }
The codebase was fully developed in Pytorch (v1.7) \cite{NEURIPS2019_9015}. All Bayesian Optimization experiments were conducted with the BoTorch package \cite{balandat2020botorch}. For molecule generation experiments, we used the rdkit (https://github.com/rdkit/rdkit) and guacamol packages (https://github.com/BenevolentAI/guacamol). 
In addition, we have made use of the code repositories listed in Table~\ref{Appendix_Code_repos} when developping our approach and conducting experiments.

\begin{table}[h]
\setlength\belowcaptionskip{0.5pt}
\begin{center}
\caption{\textbf{Code repositories used}}
\resizebox{\textwidth}{!}{
\begin{tabular}{ll}
\toprule
\textbf{Setting} &  \textbf{Data source} \\
\toprule
Grammar VAE & \url{https://github.com/mkusner/grammarVAE} \\
JT-VAE (official repo) & \url{https://github.com/wengong-jin/icml18-jtnn} \\
JT-VAE (Python 3 implementation)& \url{https://github.com/Bibyutatsu/FastJTNNpy3} \\
Quality filters & \url{https://github.com/PatWalters/rd_filters} \\
\bottomrule
\end{tabular}
}
\label{Appendix_Code_repos}
\end{center}
\end{table}
\vspace{-2mm}
\textbf{Compute resources. } All experiments were carried out with a single GPU (Titan RTX). We summarize compute usage for the different experiments in Table~\ref{Appendix_Compute_resources}.

\begin{table}[h]
\begin{center}
\caption{\textbf{Compute usage per experiment summary}}
\resizebox{\textwidth}{!}{
\begin{tabular}{lcccc}
\toprule
 &   \multicolumn{4}{c}{\textbf{Avg. compute time per iteration (GPU hrs)}} \\
\textbf{Experiment} & \textbf{Digit generation} & \textbf{Arithmetic expression} & \textbf{CVAE} & \textbf{JTVAE}\\
\toprule
Model training & <0.5 & 0.5 & 12 & 50 \\
Decoder uncertainty histograms & 2 & 2 & 3 & 40 \\
Bayesian optimization & <0.5 & <0.5 & 2 & 14 \\
Gradient ascent & <0.5 & <0.5 & 1 & 5 \\
\bottomrule
\end{tabular}
}
\label{Appendix_Compute_resources}
\end{center}
\end{table}

\vspace{-2mm}
\textbf{Data sources. } We list the raw data sources used for all experiments in Table~\ref{Appendix_Data_sources}.

\vspace{-1mm}
\begin{table}[h]
\setlength\belowcaptionskip{1pt}
\begin{center}
\caption{\textbf{Data sources summary}}
\resizebox{\textwidth}{!}{
\begin{tabular}{ll}
\toprule
\textbf{Setting} &  \textbf{Data source} \\
\toprule
Digit generation & \url{http://yann.lecun.com/exdb/mnist/} \\
Arithmetic expression & \url{https://github.com/mkusner/grammarVAE/tree/master/data} \\
Molecular generation & \url{https://github.com/aspuru-guzik-group/chemical_vae/tree/master/models/zinc} \\
\bottomrule
\end{tabular}
}
\label{Appendix_Data_sources}
\end{center}
\end{table}

\end{document}